%% file: 0main.tex
\documentclass[sigconf]{acmart}
\usepackage[ruled,linesnumbered]{algorithm2e}
\usepackage{algorithmic}
\usepackage{amsmath}
\usepackage{subcaption} 
\usepackage{multirow}
\usepackage{placeins}
\usepackage{color}
\DeclareMathOperator*{\argmax}{arg\,max}
\newtheorem{definition}{Definition}
\newtheorem{problem}{Problem}
\newtheorem{conjecture}{Conjecture}

\AtBeginDocument{%
  \providecommand\BibTeX{{%
    \normalfont B\kern-0.5em{\scshape i\kern-0.25em b}\kern-0.8em\TeX}}}

\copyrightyear{2022}
\acmYear{2022}
\setcopyright{acmcopyright}
\acmConference[CIKM '22]{Proceedings of the 31st ACM International Conference on Information and Knowledge Management}{October 17--21, 2022}{Atlanta, GA, USA}
\acmBooktitle{Proceedings of the 31st ACM International Conference on Information and Knowledge Management (CIKM '22), October 17--21, 2022, Atlanta, GA, USA}
\acmPrice{15.00}
\acmDOI{10.1145/3511808.3557437}
\acmISBN{978-1-4503-9236-5/22/10}

\begin{document}
\title{Robust Node Classification on Graphs: Jointly from Bayesian Label Transition and Topology-based Label Propagation}

\author{Jun Zhuang}
\affiliation{%
 \institution{Indiana University-Purdue University Indianapolis}
  \city{Indianapolis}
  \state{IN}
  \country{USA}
}
\email{junz@iu.edu}

\author{Mohammad Al Hasan}
\affiliation{%
  \institution{Indiana University-Purdue University Indianapolis}
  \city{Indianapolis}
  \state{IN}
  \country{USA}
}
\email{alhasan@iupui.edu}

\begin{abstract}
Node classification using Graph Neural Networks (GNNs) has been widely applied in various real-world scenarios. However, in recent years, compelling evidence emerges that the performance of GNN-based node classification may deteriorate substantially by topological perturbation, such as random connections or adversarial attacks. Various solutions, such as topological denoising methods and mechanism design methods, have been proposed to develop robust GNN-based node classifiers but none of these works can fully address the problems related to topological perturbations. Recently, the Bayesian label transition model is proposed to tackle this issue but its slow convergence may lead to inferior performance. In this work, we propose a new label inference model, namely LInDT, which integrates both Bayesian label transition and topology-based label propagation for improving the robustness of GNNs against topological perturbations. LInDT is superior to existing label transition methods as it improves the label prediction of uncertain nodes by utilizing neighborhood-based label propagation leading to better convergence of label inference. Besides, LIndT adopts asymmetric Dirichlet distribution as a prior, which also helps it to improve label inference. Extensive experiments on five graph datasets demonstrate the superiority of LInDT for GNN-based node classification under three scenarios of topological perturbations.
\end{abstract}


\maketitle

\input{1intro.tex}

\input{3method.tex}

\input{4exp.tex}

\input{2rewk.tex}

\section{Conclusion}
\label{sec:con}
In this work, we aim to improve the robustness of a GNN-based node classifier against topological perturbations given that the node classifier is trained with manual-annotated labels. To achieve this goal, we propose a new label inference model, namely LInDT, that integrates both Bayesian label transition and topology-based label propagation. Extensive experiments demonstrate the following conclusions. Firstly, GNN-based node classification can benefit from our model under three scenarios of topological perturbations. Furthermore, our model converges faster, achieves higher accuracy on node classification while maintaining normalized entropy at a low level, and surpasses eight popular competing methods across five public graph datasets. Lastly, adopting asymmetric Dirichlet distributions as a prior can contribute to label inference.

\begin{acks}
Our work is supported by NSF with grant number IIS-1909916.
\end{acks}

\appendix

\section{Implementation}
\subsection{Hardware and Software}
We conduct the experiments using Python 3.8 and PyTorch 1.7. on Ubuntu 18.04.5 LTS with Intel(R) Xeon(R) Gold 6258R CPU \@ 2.70 GHz and NVIDIA Tesla V100 PCIe 16GB GPU.

\subsection{Model Architecture and Hyper-parameters}
Our proposed model can be applied on top of GNNs. In this paper, we choose a two-layer GCN with 200 hidden units and a ReLU activation function. The GCN is trained by Adam optimizer with $1 \times 10^{-3}$ learning rate and converged within 200 epochs on all datasets.

\subsection{Hyper-parameters of Competing Methods}
We present the hyper-parameters of competing methods for reproducibility purposes. All models are trained by Adam optimizer. \\
{\bf GNN-Jaccard}~\cite{wu2019adversarial}: The similarity threshold is 0.01. The hidden units are 16. The dropout rate is 0.5. The training epochs are 300. \\
{\bf GNN-SVD}~\cite{entezari2020all}: The number of singular values is 15. Hidden units are 16. The dropout rate is 0.5. The training epochs are 300. \\
{\bf DropEdge}~\cite{rong2019dropedge}: We choose GCN as the base model with 1 base block layer and train this model with 300 epochs. Rest of the parameters are mentioned in Tab.~\ref{tab:dropedge_para}. \\
{\bf GRAND}~\cite{feng2020graph}: The model is trained with 200 epochs. The hidden units, drop node rate, and L2 weight decay is 32, 0.5, and $5 \times 10^{-4}$, respectively. The rest of parameters are reported in Tab.~\ref{tab:grand_para}. \\
{\bf RGCN}~\cite{zhu2019robust}: We set up $\gamma$ as 1, $\beta_1$ and $\beta_2$ as $5 \times 10^{-4}$ on all datasets. The hidden units for each dataset are 64, 64, 128, 1024, and 1024, respectively. The dropout rate is 0.6. The learning rate is 0.01. The training epochs are 400. \\
{\bf ProGNN}~\cite{jin2020graph}: $\alpha$, $\beta$, $\gamma$, and $\lambda$ is $5 \times 10^{-4}$, 1.5, 1.0, and 0.0, respectively. The hidden units are 16. The dropout rate is 0.5. The learning rate is 0.01. Weight decay is $5 \times 10^{-4}$. The training epochs are 100. \\
{\bf GDC}~\cite{hasanzadeh2020bayesian}: The number of blocks and layers is 2 and 4, respectively. The hidden units are 32. The dropout rate is 0.5. Both learning rate and weight decay are $5 \times 10^{-3}$. The training epochs are 400. \\
{\bf MC Dropout}~\cite{gal2016dropout}: We use the same GCN architecture and training setting as ours except for the optimal dropout rate for each dataset as 0.7, 0.6, 0.9, 0.6, and 0.8, respectively. Note that MC Dropout will apply dropout in test data. \\

\begin{table}[t]
\small
\centering
\setlength{\tabcolsep}{2pt}
\caption{Hyper-parameters of DropEdge in this paper.}
\label{tab:dropedge_para}
\begin{tabular}{cccccc}
 \toprule
  \textbf{Hyper-parameters} & \textbf{Cora} & \textbf{Citeseer} & \textbf{Pubmed} & \textbf{AMZcobuy} & \textbf{Coauthor} \\
 \midrule
    Hidden units & 128 & 128 & 128 & 256 & 128 \\
    Dropout rate & 0.8 & 0.8 & 0.5 & 0.5 & 0.5 \\
    Learning rate & 0.01 & 9e-3 & 0.01 & 0.01 & 0.01 \\
    Weight decay & 5e-3 & 1e-3 & 1e-3 & 0.01 & 1e-3 \\
    Use BN & $\times$ & $\times$ & $\times$ & $\checkmark$ & $\times$ \\
 \bottomrule
\end{tabular}
\end{table}

\begin{table}[t]
\footnotesize
\centering
\setlength{\tabcolsep}{2pt}
\caption{Hyper-parameters of GRAND in this paper.}
\label{tab:grand_para}
\begin{tabular}{cccccc}
 \toprule
  \textbf{Hyper-parameters} & \textbf{Cora} & \textbf{Citeseer} & \textbf{Pubmed} & \textbf{AMZcobuy} & \textbf{Coauthor} \\
 \midrule
    Propagation step & 8 & 2 & 5 & 5 & 5 \\
    Data augmentation times & 4 & 2 & 4 & 3 & 3 \\
    CR loss coefficient & 1.0 & 0.7 & 1.0 & 0.9 & 0.9 \\
    Sharpening temperature & 0.5 & 0.3 & 0.2 & 0.4 & 0.4 \\ 
    Learning rate & 0.01 & 0.01 & 0.2 & 0.2 & 0.2 \\
    Early stopping patience & 200 & 200 & 100 & 100 & 100 \\
    Input dropout & 0.5 & 0.0 & 0.6 & 0.6 & 0.6 \\
    Hidden dropout & 0.5 & 0.2 & 0.8 & 0.5 & 0.5 \\
    Use BN & $\times$ & $\times$ & $\checkmark$ & $\checkmark$ & $\checkmark$ \\
 \bottomrule
\end{tabular}
\end{table} 

\bibliographystyle{ACM-Reference-Format}
\bibliography{2reference}
\clearpage

\end{document}

%% file: 1intro.tex
\section{Introduction}
\label{sec:intro}
Among the various machine learning tasks on the graph data, node classification is probably the most popular with numerous real-world applications, such as, user profiling in online social networks (OSNs) \cite{chen2019semi, zhuang2022deperturbation, zhuang2022does}, community detection~\cite{chen2017supervised, breuer2020friend}, expert finding in community-based question answering~\cite{zhao2016expert, fang2016community}, recommendation systems~\cite{ying2018graph, fan2019graph, gao2022graph}, and epidemiology study~\cite{la2020epidemiological, hsieh2020drug}. While many traditional approaches have been used for solving a node classification task~\cite{mountain1997regional, bhagat2011node, tang2016node, li2019learning}, in this deep learning era various graph neural network models (GNNs)~\cite{bruna2013spectral, defferrard2016convolutional, kipf2016semi, hamilton2017inductive, du2017topology} have become popular for solving this task. The enormous appeal of GNN models for node classification is due to their ability to learn effective node-level representation by performing locality-based non-linear aggregation jointly over node attributes and adjacency information leading to superior prediction performance. 

However, in recent years, compelling evidence emerges that the performance of GNN-based node classification may deteriorate substantially by changes in a graph structure~\cite{sun2018adversarial, jin2020adversarial}, which raises concern regarding the application of GNNs in various real-life scenarios. For instance, in an online social network, non-malicious selfish users may randomly connect with many other nodes for promoting their business activities~\cite{zhuang2022deperturbation, hahn2020random}; such random connections exacerbate over-smoothing~\cite{liu2020towards, hasanzadeh2020bayesian}, a well-known cause of the poor performance of GNNs. On the other extreme, the message-passing mechanism of GNNs performs poorly on nodes with sparse connection \cite{tam2020fiedler, ye2021sparse}, such as the nodes representing new users of OSNs, which lack sufficient links (edges) or profile information (feature). Third, GNNs may be vulnerable to adversarial attacks \cite{zugner2018adversarial, dai2018adversarial, zugner2019adversarial, wang2018attack, dai2022targeted}. Such attacks may significantly deteriorate the performance of node classifications with slight and unnoticeable modification of graph structures and node features. In this paper, we refer to all such scenarios as {\bf topological perturbations}.

Many recent works have been proposed to develop robust GNN-based node classification models against topological perturbations. Most of these works can be divided into two categories: the topological denoising methods, which design methodologies for denoising perturbed graphs in a pre-processing stage \cite{wu2019adversarial, entezari2020all} or in the training stage \cite{xu2020unsupervised, luo2021learning, rong2019dropedge, tang2020transferring, zheng2020robust}, and the mechanism design methods, which make GCN's prediction robust by message-passing-based aggregation over the nodes~\cite{seddik2021node, chen2021understanding, jin2021node, zhang2020gnnguard, feng2020graph} or through regularization~\cite{chang2021not, liu2021elastic, regol2022node, xu2021speedup, jin2020graph, hasanzadeh2020bayesian, xu2020towards, zhu2019robust}. Nevertheless, none of these works can fully address the topological perturbations. For one thing, topological denoising methods may fail to prune suspicious edges when the graph structure is sparse. For another, mechanism designing methods sometimes highly depend on heuristics explorations of topological connections among nodes. The mechanism may perform worse when the graph structure is under severe perturbations.

In some recent works~\cite{zhuang2022deperturbation, zhuang2022defending}, the authors tackle the topological perturbation issues by developing a Bayesian label transition mechanism, which fixes the poor performance of GNNs by post-processing the prediction; specifically, their approaches learn a label transition matrix through Bayesian inferences, which fixes the erroneous prediction of GNNs by replacing the GNN inferred labels to a different label. A key benefit of post-processing is that it is effective for dynamic network scenarios, in which the network structure and node features of test data may be different than the train data. However, Bayesian label transition~\cite{zhuang2022defending} suffers slow convergence as the posterior distribution from which the Gibbs sampling samples the node labels adapt incrementally over a sequence of label-propagation followed by GNN re-training iterations. Slow convergence also leads to inferior performance because the labels on which the GNN is re-trained may not be accurate as the Bayesian label transition may have not yet been converged.

Although label transition is a relatively new idea in the node classification task, label propagation (LP) has been widely used in the semi-supervised learning paradigm \cite{zhu2002learning, wang2007label, speriosu2011twitter, wang2013dynamic, karasuyama2013multiple, ugander2013balanced, gong2016label, iscen2019label, huang2020combining, wang2020unifying}. Recently, Huang et al.~\cite{huang2020combining} demonstrated that graph semi-supervised learning using LP can exceed or match the performance of some variants of GNNs. Zhu and Ghahramani~\cite{zhu2002learning} have shown that label propagation is similar to mean-field approximation, which has fast convergence when the number of instances is large. The main assumption of LP is that closer data points tend to have similar class labels---this assumption is analogous to the graph Homophily assumption, which denotes that adjacent vertices have similar labels. Specifically, on noisy and perturbed graphs, the label prediction on a given (or a targeted) node can be affected severely, however, the prediction of other neighboring nodes may still be sufficiently good, and by the virtue of homophily, those predictions can help recovering the true label of the affected node. Inspired by these observations, in this work,  we improve the robustness of GNN-based node classification through an integrated framework combining Bayesian label transition and graph topology-based label propagation. We name our proposed model {\bf LInDT}~\footnote{LInDT is built by taking the bold letters of ``{\bf L}abel {\bf In}ference using {\bf D}irichlet and {\bf T}opological sampling'', and its pronunciation is similar to the popular chocolate brand name. Source code is available at \textbf{https://github.com/junzhuang-code/LInDT}}.

The main advantage of LInDT over a label-transition-based method, such as GraphSS~\cite{zhuang2022defending}, is that the former effectively utilizes LP when the sampling distribution of Bayesian label transition is uncertain. This immediately improves the label convergence and subsequently improves the label prediction performance. Besides, LInDT adopts a better Bayesian prior than GraphSS; specifically, LInDT follows an informative asymmetric Dirichlet distribution, but GraphSS follows a symmetric Dirichlet distribution, which allows LInDT to take advantage of the most up-to-date label distribution, leading to better performance. We evaluate LInDT's performance under random perturbations, information sparsity, and adversarial attacks scenarios, across five public graph datasets. The experimental results verify the following facts: First, the GNN-based node classifier benefits strongly from LInDT's proposed label inference as post-processing; Second, LInDT converges faster than a pure Bayesian label transition which only uses Gibbs based posterior sampling; Third, LInDT outperforms eight popular competing methods. We also perform an ablation study to verify the effectiveness of LInDT's design choices. For example, our experiment validates that asymmetric Dirichlet distributions of LInDT help better label inference using Bayesian label transition.

Overall, we summarize our contributions in this paper as follows:
\begin{itemize}
  \item We propose a new label inference mechanism that can infer node labels jointly from a Bayesian and label propagation framework. Our model can converge faster than the Bayesian label transition using Gibbs sampling.
  \item We release the constraint of symmetric Dirichlet distributions on the Bayesian label transition and verify that dynamically updating the $\alpha$ vector (Dirichlet prior) can help better label inference.
  \item Extensive experiments demonstrate that our model achieves robust node classification on three topological perturbation scenarios and outperforms eight popular defending models across five public graph datasets.
\end{itemize}

%% file: 3method.tex
\section{Methodology}
\label{sec:method}
In this section, we first briefly introduce the node classification using GNNs. We then describe the basic idea of Bayesian label transition. Besides, we discuss asymmetric Dirichlet distributions and introduce our proposed model. Lastly, we present the pseudo-code and time complexity of our model.

\subsection{GNN-based Node Classification}
\label{subsec:gnn}
An undirected attributed graph is denoted as $\mathcal{G}$ = $(\mathcal{V}$, $\mathcal{E})$, where $\mathcal{V}$ = $\{ v_{1}, v_{2}, ..., v_{N} \}$ is the set of vertices, $N$ is the number of vertices in $\mathcal{G}$, and $\mathcal{E} \subseteq \mathcal{V} \times \mathcal{V}$ is the set of edges between vertices. We denote $\mathbf{A} \in \mathbb{R}^{N \times N}$ as symmetric adjacency matrix and $\mathbf{X} \in \mathbb{R}^{N \times d}$ as the feature matrix, where $d$ is the number of features for each vertex. We choose the most representative variant of GNNs, GCN \cite{kipf2016semi}, as the node classifier $\textit{f}_{\theta}$ in this work. The layer-wise propagation of GCN is formulated as follows:
\begin{equation}
\mathbf{H}^{(l+1)} = \sigma \left( \mathbf{\tilde{D}}^{-\frac{1}{2}} \mathbf{\tilde{A}} \mathbf{\tilde{D}}^{-\frac{1}{2}} \mathbf{H}^{(l)} \mathbf{W}^{(l)} \right)
\label{eqn:gcn}
\end{equation}
In Eq. \ref{eqn:gcn}, $\mathbf{\tilde{A}} = \mathbf{A} + I_{N}$, $\mathbf{\tilde{D}} = \mathbf{D} + I_{N}$, where $I_{N}$ is the identity matrix and $\mathbf{D}_{i,i} = \sum_{j} \mathbf{A}_{i,j}$ is the diagonal degree matrix. $\mathbf{H}^{(l)} \in \mathbb{R}^{N \times d}$ is the nodes hidden representation in the $l$-th layer, where $\mathbf{H}^{(0)} = \mathbf{X}$. $\mathbf{W}^{(l)}$ is the weight matrix in the $l$-th layer. $\sigma(\cdot)$ denotes a non-linear activation function, such as ReLU.

In general node classification tasks, a GNN-based node classifier $\textit{f}_{\theta}$ takes both $\mathbf{A}$ and $\mathbf{X}$ as inputs and is trained with ground-truth labels in train data. To obtain robust classification, some studies attempt to train the model using noisy labels as a regularization scheme~\cite{tanno2019learning, li2020dividemix}. In this study, we borrow this idea and train the GNN node classifier, $\textit{f}_{\theta}$, after assigning a uniform random label to a percentage (in this work we use 10\%) of the train nodes. We denote these {\bf noisy labels} as $\mathcal{Y} \in \mathbb{R}^{N \times 1}$. The higher the percentage of noise ratio, the stronger the regularization. On the other hand, we denote the ground truth labels (a.k.a. {\bf latent labels}) as $\mathcal{Z} \in \mathbb{R}^{N \times 1}$, which are unobserved (to the GNN), yet we want the GNN to be able to predict the unobserved label. So, the model's performance is optimized to predict $\mathcal{Z}$ by using a validation set. It may seem counter-intuitive to train a model with noisy labels and still expect it to perform well on the ground-truth label; however, this is feasible because both $\mathcal{Z}$ and $\mathcal{Y}$ take values from the same closed category set. Also, there is a strong correlation between $\mathcal{Y}$ and $\mathcal{Z}$ as only a fraction of the node's labels have been altered. But, as GNN is trained on the noisy labels, it becomes noise-resilient and robust, and performs well when the network is perturbed.

\subsection{Bayesian Label Transition}
\label{subsec:blt}

\begin{figure}[h]
  \centering
  \includegraphics[width=\linewidth]{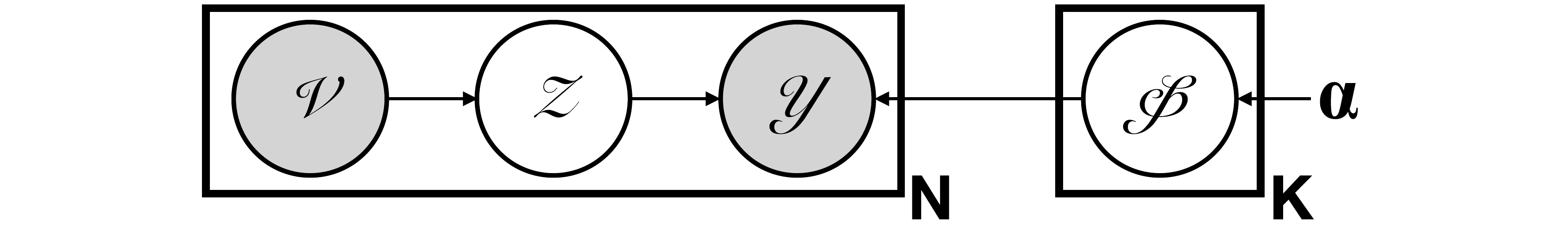}
  \caption{The diagram of Bayesian label transition. $\mathcal{V}$, $\mathcal{Z}$, and $\mathcal{Y}$ denote $N$ copies of nodes, latent labels, and noisy labels, respectively. $\phi$ denotes the $K$-class Dirichlet-based conditional label transition matrix parameterized by $\alpha$.}
\label{fig:fig_lt}
\end{figure}

A GNN suffers performance loss when applied in the test data, which undergoes various topological perturbations. To overcome this, recent works have applied Bayesian label transition to fix erroneous prediction of GNN by learning a label transition matrix on the fly (from the test data).
We explain Bayesian label transition by utilizing the plate diagram in Fig. \ref{fig:fig_lt}. The unobserved latent labels ($\mathcal{Z}$) depend on the nodes, whereas the observed noisy labels ($\mathcal{Y}$) depend on both $\mathcal{Z}$ and the conditional label transition matrix, $\phi$, which is parameterized by $\alpha$. Both $\mathcal{Z}$ and $\mathcal{Y}$ are sampled from $Categorical$ distributions of $\mathcal{V}$, the predictions from the trained node classifier. $\phi$ = $[\phi_{1}, \phi_{2}, …, \phi_{K}]^{T}$ $\in \mathbb{R}^{K \times K}$ consists of K transition vectors. The $k$-th transition vector $\phi_{k}$ is sampled from corresponding Dirichlet distribution $Dirichlet(\alpha_k)$, where $\alpha_k$ is the $k$-th element of the $\alpha$ vector. Based on this dependency, the posterior of $\mathcal{Z}$, which is conditioned on the nodes $\mathcal{V}$, the noisy labels $\mathcal{Y}$, and the Dirichlet parameter $\alpha$, can be formulated as follows:
\begin{equation}
\textit{P} \left( \mathcal{Z} \mid \mathcal{V}, \mathcal{Y} ; \alpha \right) = \textit{P} \left( \mathcal{Z} \mid \mathcal{V}, \mathcal{Y}, \phi \right) \textit{P} \left( \phi ; \alpha \right) \\
\label{eqn:dependency}
\end{equation}
However, sampling from such a posterior distribution is difficult. GraphSS~\cite{zhuang2022defending} applies Gibbs sampling to approximate this posterior, so that {\bf inferred labels} $\mathcal{\bar{Z}} \in \mathbb{R}^{N \times 1}$ matches to the latent labels ($\mathcal{Z})$ as identically as possible. 

In this study, we assume the test graph is perturbed through various structural changes and further develop the Bayesian label transition model to improve the robustness of GNN-based node classifiers against topological perturbations. Before summarizing the problem we aim to solve in this work, we first clarify that noisy labels in this work include both {\bf manual-annotated labels} $\mathcal{Y}_m \in \mathbb{R}^{N_{train} \times 1}$ and {\bf auto-generated labels} $\mathcal{Y}_a \in \mathbb{R}^{N_{test} \times 1}$, where $N_{train}$ and $N_{test}$ denote the number of nodes on the train and test graphs, respectively.
More specifically, we first train GNN with $\mathcal{Y}_m$ on the unperturbed train graph to generate $\mathcal{Y}_a$. The prediction of the test nodes may perform worse when the test graph is under topological perturbations. At this point, we employ our proposed label inference model to iteratively recover the perturbed predictions by $\mathcal{Y}_a$ with several epochs of transition. After the label transition is converged, we expect that the inferred labels will be as identical as the latent labels. Overall, the above-mentioned problem could be formally described as follows:
\begin{problem}
Given a graph $\mathcal{G} = (\mathbf{A}, \mathbf{X})$ and the manual-annotated labels $\mathcal{Y}_m$, which are used for training a GNN-based node classifier $\textit{f}_{\theta}$ on unperturbed $\mathcal{G}_{train}$, we aim to develop a label inference model to improve the robustness of the node classifier $\textit{f}_{\theta}$ on perturbed $\mathcal{G}_{test}$ by approximating the inferred labels $\mathcal{\bar{Z}}$ to the latent labels $\mathcal{Z}$ as identical as possible.
\end{problem}

\subsection{Asymmetric Dirichlet Distributions}
\label{subsec:ads}
\begin{figure}[h]
  \centering
  \includegraphics[width=\linewidth]{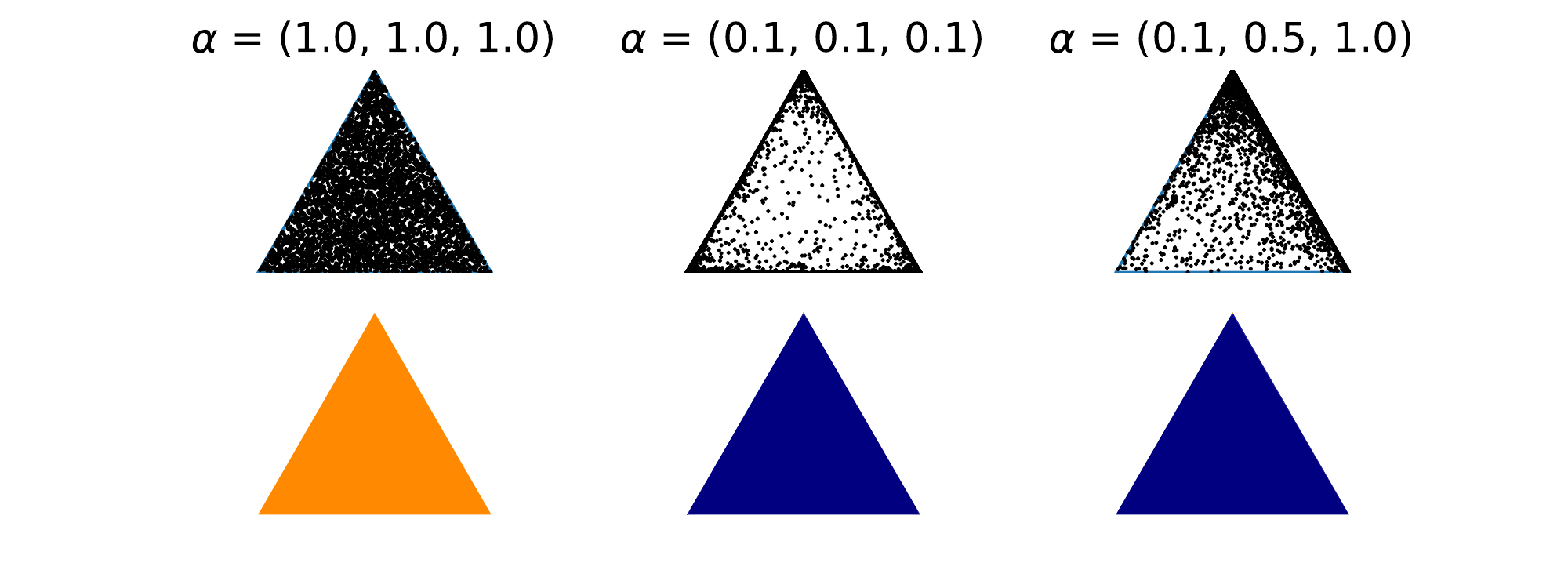}
  \caption{Toy examples of Dirichlet distributions.}
\label{fig:fig_dirichlet}
\end{figure}

In Eq. \ref{eqn:dependency}, GraphSS~\cite{zhuang2022defending} assumes the Dirichlet distribution is symmetric, i.e., the parameter vector $\alpha$ has the same value, 1.0, to all classes. In this study, we release this constraint and investigate whether adopting asymmetric Dirichlet distributions as a prior can contribute to the label transition.
Fig. \ref{fig:fig_dirichlet} visualizes the Dirichlet distributions by equilateral triangles. The first triangle represents the symmetric Dirichlet distribution. The second triangle shows that data points tend to move to the corner with the same probability. The third triangle presents that data points have a higher probability to be assigned to the class whose $\alpha$ value is higher.
The intuition is that the $i^{th}$ inferred label $\mathbf{\bar{z}}^{t}_{i}$ may have higher probability to be transited to the $k^{th}$ class in the $t^{th}$ transition when $\alpha^{t}_{k}$ is higher. Following this intuition, we dynamically update the $k^{th}$-class $\alpha$ value, $\alpha^{t}_{k}$, in the $t^{th}$ transition as follows:
\begin{equation}
\alpha^{t}_{k} = \alpha^{t-1}_{k} \frac{\sum_{i=1}^{N} {I(\mathbf{\bar{z}}^{t}_{i}  = k)}}{\sum_{i=1}^{N} {I(\mathbf{\bar{z}}^{t-1}_{i} = k)}}
\label{eqn:dynamic_alpha}
\end{equation}
where $I(\cdot)$ denotes an indicator function.

\subsection{Label Inference Jointly from Bayesian Label Transition and Topology-based Label Propagation}
\label{subsec:lindt}
GraphSS~\cite{zhuang2022defending} applies Gibbs sampling to recover the multinominal distribution of nodes on perturbed graphs. However, such sampling may converge slowly and thus leads to inferior inference. This drawback can be mitigated by considering the fact that the majority of real-life networks exhibit homophily property, i.e., labels of the adjacent nodes are similar. So, if the inferred label of a node during label transition is uncertain, we can sample its label from the labels of adjacent nodes. This idea is inspired by well-known Label propagation algorithms~\cite{wang2013dynamic}. Thus, LInDT integrates both Bayesian label transition and topology-based label propagation. More specifically, for each node on $\mathcal{G}_{test}$, our model first infers the label from Bayesian label transition and then substitutes this label using topology-based label propagation when this inferred label is uncertain. We define the node label uncertainty during the inference as follows:

\begin{definition}
In the $t^{th}$ transition, an inferred label $\mathbf{\bar{z}}^{t}$ is uncertain, such that $\mathbf{\bar{z}}^{t} \neq \mathbf{\bar{z}}^{t-1}$ or $\mathbf{\bar{z}}^{t} \neq \mathbf{y}$.
\end{definition}
Note that $\mathbf{\bar{z}}^{t} \neq \mathbf{\bar{z}}^{t-1}$ will be happened when the inference is not converged yet, whereas $\mathbf{\bar{z}}^{t} \neq \mathbf{y}$ indicates that the inferred labels $\mathcal{\bar{Z}}$ still deviates from the auto-generated labels $\mathcal{Y}_a$.
The intuition of our model is that unnoticeable perturbations may severely deteriorate the predictions of GNN-based node classifiers, but fortunately, such perturbations only alter a small number of topological structures. Based on the homophily assumption of the graph structure, i.e., intra-class nodes tend to connect with each other, sampling labels with topology-based label propagation can decrease the node label uncertainty by utilizing topological information of nodes. Note that we present the edge homophily ratio ({\bf EHR} $\in[0, 1]$)~\cite{zhu2020beyond}, the percentage of intra-class edges, in Tab. \ref{table:dataset} to verify the homophily property in the graph datasets. Higher EHR indicates stronger homophily property.

\begin{algorithm}
\DontPrintSemicolon
\KwIn{Categorical distribution $\overline{\textit{P}} \left(\mathcal{\bar{Z}}^{t-1} \mid \mathcal{V} \right)$, Transition matrix $\phi^{t-1}$, Topology-based label sampler.}
\For{$i \gets 0$  $\textbf{to}$  $N$} {
    $\mathbf{\bar{z}}^{t}_{i} \sim \argmax \overline{\textit{P}} \left(\mathbf{\bar{z}}^{t-1}_{i} \mid v_{i} \right) \phi^{t-1}$; \\
    \If{$\mathbf{\bar{z}}^{t}_{i}$ is uncertain} {
        Update $\mathbf{\bar{z}}^{t}_{i}$ with the topology-based label sampler;
    }
}
\Return{Inferred labels $\mathcal{\bar{Z}}^{t}$ in the $t^{th}$ transition.}
\caption{{\sc Label} Sampling Jointly from Bayesian Label Transition and Topology-based Label Propagation}
\label{algo:lsdt}
\end{algorithm}

We present how our model samples labels jointly from Bayesian Label Transition and Topology-based Label Propagation in Algo. \ref{algo:lsdt}. In the $t^{th}$ transition, we first sample a label $\mathbf{\bar{z}}^{t}_{i}$ of the $i^{th}$ node from the categorical distribution $\overline{\textit{P}} (\mathbf{\bar{z}}^{t-1}_{i} \mid v_{i} ) \in \mathbb{R}^{N \times K}$ updated with a given label transition matrix $\phi^{t-1}$ in the $(t-1)^{th}$ transition (\textbf{Line 2}). We then update $\mathbf{\bar{z}}^{t}_{i}$ with a given topology-based label sampler when this inferred label is uncertain (\textbf{Line 3-4}). In the end, we obtain the inferred labels in the $t^{th}$ transition.

\begin{figure}[h]
  \centering
  \includegraphics[width=\linewidth]{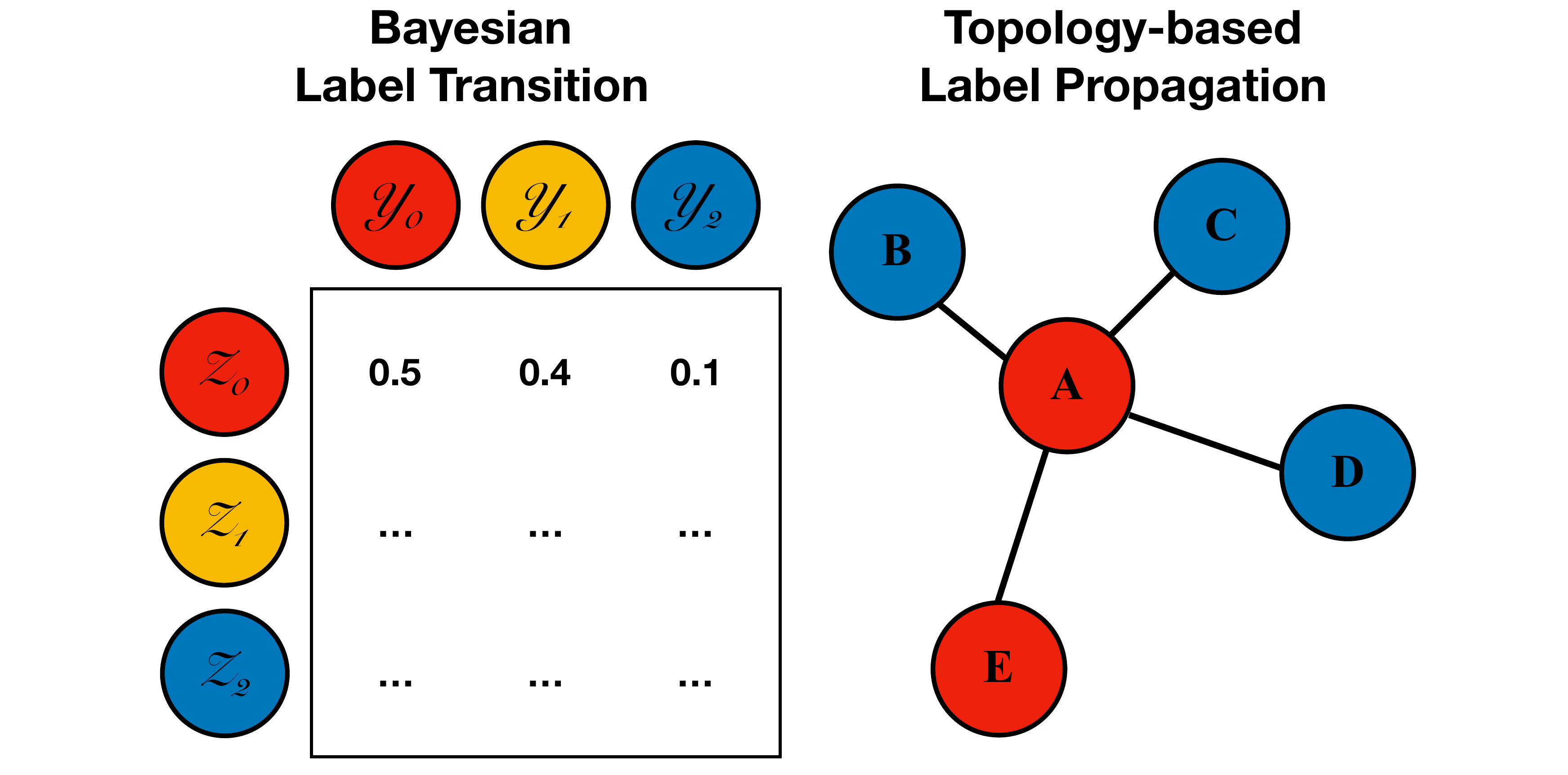}
  \caption{Toy example of our proposed sampling method.}
\label{fig:fig_space}
\end{figure}

Besides Algo. \ref{algo:lsdt}, we further use a toy example to explain our proposed sampling method in Fig. \ref{fig:fig_space}. In the $t^{th}$ transition, our model first samples the inferred label of node A as the red class via Bayesian label transition. However, this inferred label is different from the auto-generated label of node A; let's say it's blue. In this case, we say that this node label is uncertain. Based on the assumption of homophily graphs and unnoticeable perturbations, i.e., the labels of majority neighbor nodes are predicted correctly in $\mathcal{Y}_a$, we then apply the topology-based label sampler to substitute the label, i.e., replacing it from red to blue based on the topological information of node A.

\begin{figure}[h]
  \centering
  \includegraphics[width=\linewidth]{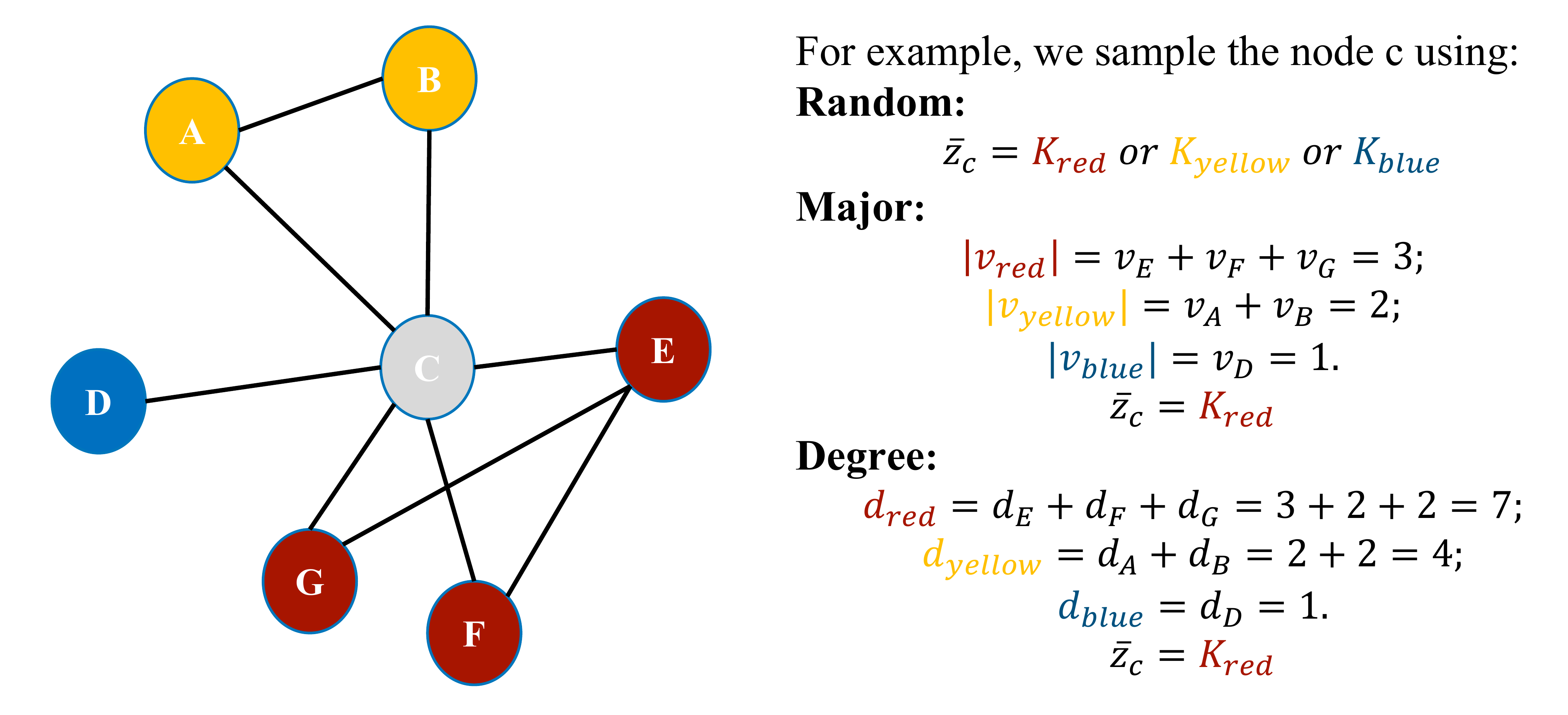}
  \caption{Toy examples of our three proposed samplers. We use red, yellow, and blue color to represent three classes.}
\label{fig:fig_ts}
\end{figure}

Based on the above-mentioned intuition, we investigate three types of topology-based label samplers, random sampler ({\bf Random}), majority sampler ({\bf Major}), and degree-weighted sampler ({\bf Degree}). All of these samplers sample a label from 1-hop neighborhood of the current node. We denote the set of classes from the 1-hop neighborhood as $K_{nei}$ and the number of neighbour nodes as $N_{nei}$. We highlight the distinctions of these candidate samplers as follows and further explain them with toy examples in Fig. \ref{fig:fig_ts}.
\begin{itemize}
\item Random sampler randomly samples a label. In the $t^{th}$ transition, the probability that the $i^{th}$ inferred label $\mathbf{\bar{z}}^{t}_{i}$ is equal to $k^{th}$ class can be formulated as Eq. \ref{eqn:random}.
\begin{equation}
\textit{P} \left(\mathbf{\bar{z}}^{t}_{i} = k \mid v_{i} \right)
= \frac{\sum_{i=1}^{N_{nei}} {I(\mathbf{\bar{z}}^{t}_{i} = k)}}{\sum_{i=1}^{N_{nei}} {I(\mathbf{\bar{z}}^{t}_{i} \in K_{nei})}}
\label{eqn:random}
\end{equation}
\item Majority sampler selects the majority class $k_{mj}$ as the label, such that the number of nodes in majority class $|v_{mj}| = max \{ \sum_{\mathbf{\bar{z}}_{j}=k} \mathbf{A}_{ij} : k \in K_{nei} \}$.
\item Degree-weighted sampler chooses a label from the class $k_{dw}$, such that sum of the neighbor nodes' total degrees in $k_{dw}$ is maximum. This maximum sum can be expressed as $max \{ \sum_{\mathbf{\bar{z}}_{i}=k} d_{i} : k \in K_{nei} \}$, where $d_{i} = \sum_{j} \mathbf{A}_{ij}$ is the total degree of the $i^{th}$ node.
\end{itemize}

{\bf Analysis of Convergence.}
To analyze the convergence of our proposed sampling method, we discuss two conjectures as follows.

\begin{conjecture}
Given a large enough iterations of transition $T$, the transition of inferred labels $\mathcal{\bar{Z}}$ will eventually converge to $\mathcal{\bar{Z}}^{\pi}$.
\end{conjecture}
{\bf Analysis}: We formally state the transition as follows:
\begin{equation}
\small
\mathcal{\bar{Z}^{\pi}}
= \lim_{T \to \infty} \argmax \overline{\textit{P}} \left(\mathcal{\bar{Z}}^T \mid \mathcal{V} \right)
= \lim_{T \to \infty} \argmax \overline{\textit{P}} \left(\mathcal{\bar{Z}}^{T-1} \mid \mathcal{V} \right) \phi^{T-1} \\
\label{eqn:convergency}
\end{equation}
Theoretical proof of this conjecture is difficult because both the categorical distribution $\overline{\textit{P}} \left(\mathcal{\bar{Z}} \mid \mathcal{V} \right)$ and the label transition matrix $\phi$ in each transition will be dynamically updated. Instead, we empirically examine this conjecture in the experiment. Note that the converged inferred labels $\mathcal{\bar{Z}^{\pi}}$ is not guaranteed to be the same as the latent labels $\mathcal{Z}$, which translates to the
possibility of erroneous  prediction.

\begin{conjecture}
Label sampling jointly from Bayesian Label Transition and Topology-based Label Propagation helps the inference converge with fewer iterations of transition than the Bayesian-based sampling.
\end{conjecture}
{\bf Analysis}: Our proposed sampling method will substitute the inferred labels of uncertain nodes based on the topological information. On the homophily graphs, we can get smaller Total Variation Distance between current node label distributions of inferred labels in $t^{th}$ transition $\mathcal{\bar{Z}}^t$ and the convergence distributions of inferred labels $\mathcal{\bar{Z}}^{\pi}$ as follows:
\begin{equation}
\small
\lVert \mathcal{\bar{Z}}^t_{DT} - \mathcal{\bar{Z}}^{\pi} \rVert_{TV} \leq \lVert \mathcal{\bar{Z}}^t_{Bayes} - \mathcal{\bar{Z}}^{\pi} \rVert_{TV} \\
\label{eqn:mixing_time}
\end{equation}
where $\mathcal{\bar{Z}}^t_{DT}$ denotes the inferred labels using our proposed sampling method, whereas $\mathcal{\bar{Z}}^t_{Bayes}$ denotes the inferred labels using the Bayesian-based sampling method. $\lVert \cdot \rVert_{TV}$ denotes the Total Variation Distance.

We argue that this conjecture is true as such substitutions can force the inferred labels to get closer to the convergence based on the homophily assumption, shortening the iterations of transition compared to using the Bayesian-based sampling method, e.g., Gibbs sampling. Our experiments empirically verify this conjecture.

\subsection{Pseudo-code of Our Proposed Model}
\label{subsec:algo}

\begin{algorithm}
\algsetup{linenosize=\small} \small
\DontPrintSemicolon 
\KwIn{Graph $\mathcal{G}_{train}$ and $\mathcal{G}_{test}$, which contain corresponding symmetric adjacency matrix $\mathbf{A}$ and feature matrix $\mathbf{X}$, Manual-annotated labels $\mathcal{Y}_m$ in $\mathcal{G}_{train}$, Node classifier $\textit{f}_{\theta}$, Initial $\alpha$, The number of warm-up steps $WS$, The number of transition $T$.}
Train $\textit{f}_{\theta}$ with $\mathcal{Y}_m$ on $\mathcal{G}_{train}$; \\
Generate initial categorical distribution $\overline{\textit{P}} \left(\mathcal{Z} \mid \mathcal{V} \right)$ and auto-generated labels $\mathcal{Y}_a$ by $\textit{f}_{\theta}$; \\
Compute warm-up label transition matrix $\phi'$ based on $\mathcal{G}_{train}$; \\
Define inferred labels $\mathcal{\bar{Z}}$, dynamic label transition matrix $\phi$ based on $\mathcal{G}_{test}$, and initial $\alpha$ vector; \\
\For{$t \gets 1$  $\textbf{to}$  $T$} {
  \If{$t < WS$} {
    Sample $\mathcal{\bar{Z}}^t$ with warm-up $\phi'$ by Algo. \ref{algo:lsdt}; \\
  }
  \Else{
    Sample $\mathcal{\bar{Z}}^t$ with dynamic $\phi$ by Algo. \ref{algo:lsdt}; \\
  }
  Update $\alpha$ by Eq. \ref{eqn:dynamic_alpha} and dynamic $\phi$; \\
  Retrain $\textit{f}_{\theta}$ and update $\overline{\textit{P}} \left(\mathcal{\bar{Z}}^{t} \mid \mathcal{V} \right)$; \\
}
\Return{Inferred labels $\mathcal{\bar{Z}}$ and Dynamic $\phi$;}
\caption{{\sc LInDT}'s Pseudo-code}
\label{algo:pseudocode}
\end{algorithm}

To warp up our model in Algo. \ref{algo:pseudocode}, we discuss the pseudo-code and analyze the time complexity in this section.
{\bf Training:} Our model trains the node classifier $\textit{f}_{\theta}$ on the train graph $\mathcal{G}_{train}$ at first with manual-annotated labels $\mathcal{Y}_m$ (\textbf{line 1}) and then generates initial categorical distribution $\overline{\textit{P}} \left(\mathcal{Z} \mid \mathcal{V} \right)$ and auto-generated labels $\mathcal{Y}_a$ by $\textit{f}_{\theta}$ (\textbf{line 2}).
{\bf Inference:} Before the inference, our model first computes a warm-up label transition matrix $\phi'$ by using the prediction over $\mathcal{G}_{train}$ (\textbf{line 3}). Our model then defines (creates empty spaces) the inferred labels $\mathcal{\bar{Z}}$, the dynamic label transition matrix $\phi$ based on the test graph $\mathcal{G}_{test}$, and also initializes the $\alpha$ vector with a given initial $\alpha$ value (\textbf{line 4}).
In the warm-up stage, our model samples the inferred labels in the $t^{th}$ transition with the warm-up $\phi'$, which is built with the categorical distribution of $\textit{f}_{\theta}$ and $\mathcal{Y}_m$ on $\mathcal{G}_{train}$, via corresponding topology-based label sampler using Algo. \ref{algo:lsdt} (\textbf{line 7}).
After the warm-up stage, our model applies the same sampling method with the dynamic $\phi$ (\textbf{line 10}). This dynamic $\phi$ updates in each transition with current inferred labels $\mathcal{\bar{Z}}^t$ and corresponding $\mathcal{Y}_a$. 
Simultaneously, our model dynamically updates the $\alpha$ vector by Eq. \ref{eqn:dynamic_alpha} and the dynamic $\phi$ matrix (\textbf{line 12}).
Besides, our model iteratively re-trains the node classifier with the current inferred labels $\mathcal{\bar{Z}}^t$ in the $t^{th}$ transition, and then updates the categorical distribution $\overline{\textit{P}} \left(\mathcal{\bar{Z}}^t \mid \mathcal{V} \right)$ modeled by the re-trained $\textit{f}_{\theta}$ (\textbf{line 13}).
The transition will eventually reach convergence, approximating the inferred labels to the latent labels as identical as possible. In other words, our model can help recover the original predictions by dynamic conditional label transition on perturbed graphs.

According to Algo. \ref{algo:pseudocode}, our model applies $T$ transitions in the inference. For each transition, our proposed sampling method traverses all test nodes once. Overall, the {\bf time complexity} of the inference approximates to $\mathcal{O}(T \cdot N_{test})$, where $T$ denotes the number of transition, whereas $N_{test}$ denotes the number of nodes on the test graph.

%% file: 4exp.tex
\section{Experiments}
\label{sec:exp}
In this section, we want to answer the following questions for evaluating our model.
\begin{itemize}
    \item Can node classification benefit from our model against topological perturbations?
    \item How's the convergence of our model?
    \item How's our performance compared to competing methods?
    \item Can asymmetric Dirichlet distributions contribute to the label inference?
\end{itemize}

\begin{table}[h] 
\centering
\setlength{\tabcolsep}{4.5pt}
\caption{Statistics of datasets. $\left| \mathcal{V} \right|$, $\left| \mathcal{E} \right|$, $\left| F \right|$, and $\left| C \right|$ denote the number of nodes, edges, features, and classes, respectively. Avg.D denotes the average degree of test nodes. EHR denotes the edge homophily ratio.}
\label{table:dataset}
\begin{tabular}{ccccccc}
  \toprule
    \textbf{Dataset} & {$\left| \mathcal{V} \right|$} & {$\left| \mathcal{E} \right|$} & {$\left| F \right|$} & {$\left| C \right|$} & {Avg.D} & {EHR(\%)} \\
    \midrule
    \textbf{Cora} & 2,708 & 10,556 & 1,433 & 7 & 4.99 & 81.00 \\
    \textbf{Citeseer} & 3,327 & 9,228 & 3,703 & 6 & 3.72 & 73.55 \\
    \textbf{Pubmed} & 19,717 & 88,651 & 500 & 3 & 5.50 & 80.24 \\
    \textbf{AMZcobuy} & 7,650 & 287,326 & 745 & 8 & 32.77 & 82.72 \\
    \textbf{Coauthor} & 18,333 & 327,576 & 6,805 & 15 & 10.01 & 80.81 \\
  \bottomrule
\end{tabular}
\end{table}

\begin{table*} 
\centering
\setlength{\tabcolsep}{5pt}
\caption{Examination of our model on top of GCN under three scenarios of topological perturbations across five datasets. "Original" denotes the original performance of GCN on perturbed graphs (before inference). GraphSS is our baseline model with a fixed $\alpha$ value, 1.0. "Vanilla" denotes that we dynamically update the $\alpha$ vector using Gibbs samplers. "Random", "Major", and "Degree" denote that we employ the corresponding topology-based label sampler based on our vanilla architecture instead of using Gibbs samplers. All methods are evaluated by classification accuracy (Acc.) and average normalized entropy (Ent.) on victim nodes.}
\label{table:exp1}
\begin{tabular}{c|c|cc|cc|cc|cc|cc} 
\toprule 
\multirow{2}{*}{} & \multirow{2}{*}{\parbox{1.3cm}{\centering \textbf{Methods}}} & 
  \multicolumn{2}{c|}{\textbf{Cora}} &
  \multicolumn{2}{c|}{\textbf{Citeseer}} &
  \multicolumn{2}{c|}{\textbf{Pubmed}} &
  \multicolumn{2}{c|}{\textbf{AMZcobuy}} &
  \multicolumn{2}{c}{\textbf{Coauthor}} \\
\cline{3-12}
  &  & \textbf{Acc.} & \textbf{Ent.} & \textbf{Acc.} & \textbf{Ent.} & \textbf{Acc.} & \textbf{Ent.} & \textbf{Acc.} & \textbf{Ent.} & \textbf{Acc.} & \textbf{Ent.}  \\
\midrule
\multirow{6}{*}{\textbf{rdmPert}}
& Original & 48.95 & \textcolor{gray}{\bf 9.24} & 45.92 & 62.17 & 25.13 & \textcolor{gray}{\bf 3.59} & 81.25 & 43.83 & 38.63 & 17.10 \\
& GraphSS~\cite{zhuang2022defending} & 82.61 & 18.34 & 31.33 & \textcolor{gray}{\bf 9.35} & 81.33 & 25.25 & 83.56 & 7.93 & 88.42 & 6.60 \\
& Vanilla & 83.68 & 18.34 & 56.22 & 40.35 & 82.45 & 23.86 & 84.68 & \textcolor{gray}{\bf 7.41} & 89.66 & \textcolor{gray}{\bf 5.78} \\
& Random & \textbf{84.74} & 21.49 & \textbf{71.24} & 53.71 & 83.54 & 32.41 & 91.91 & 12.61 & 90.58 & 6.86 \\
& Major & 84.21 & 22.22 & 66.95 & 66.99 & \textbf{83.64} & 32.97 & \textbf{91.93} & 12.79 & 90.59 & 7.19 \\
& Degree & 83.63 & 26.92 & 65.32 & 66.81 & 83.52 & 32.32 & \textbf{91.93} & 12.24 & \textbf{90.62} & 7.20 \\
\midrule
\multirow{6}{*}{\textbf{infoSparse}}
& Original & 71.73 & 47.87 & 62.26 & 90.13 & 76.26 & 57.65 & 90.01 & 13.48 & 86.62 & 38.34 \\
& GraphSS~\cite{zhuang2022defending} & 79.32 & \textcolor{gray}{\bf 27.96} & \textbf{69.77} & \textcolor{gray}{\bf 46.81} & 84.03 & 29.64 & 90.81 & 7.10 & 90.15 & 8.77 \\
& Vanilla & \textbf{79.48} & \textcolor{gray}{\bf 27.96} & \textbf{69.77} & \textcolor{gray}{\bf 46.81} & 84.05 & 29.64 & 91.90 & 7.25 & 91.06 & 8.54 \\
& Random & \textbf{79.48} & \textcolor{gray}{\bf 27.96} & \textbf{69.77} & \textcolor{gray}{\bf 46.81} & 84.09 & 29.74 & 91.91 & 7.26 & 91.20 & 8.71 \\
& Major & \textbf{79.48} & \textcolor{gray}{\bf 27.96} & \textbf{69.77} & \textcolor{gray}{\bf 46.81} & \textbf{84.10} & 29.56 & 91.91 & \textcolor{gray}{\bf 6.88} & 91.22 & \textcolor{gray}{\bf 8.29} \\
& Degree & \textbf{79.48} & \textcolor{gray}{\bf 27.96} & \textbf{69.77} & \textcolor{gray}{\bf 46.81} & 84.07 & \textcolor{gray}{\bf 28.38} & \textbf{91.93} & 6.89 & \textbf{91.24} & 8.63 \\
\midrule
\multirow{6}{*}{\textbf{advAttack}}
& Original & 33.86 & 1.43 & 4.31 & 37.29 & 23.55 & 11.83 & 77.17 & 49.25 & 58.69 & 36.05 \\
& GraphSS~\cite{zhuang2022defending} & 38.10 & \textcolor{gray}{\bf 1.24} & 54.31 & \textcolor{gray}{\bf 13.04} & 83.04 & \textcolor{gray}{\bf 9.89} & 81.50 & 13.55 & 74.52 & 8.35 \\
& Vanilla & 39.68 & \textcolor{gray}{\bf 1.24} & 66.38 & 13.42 & 85.13 & \textcolor{gray}{\bf 9.89} & 82.61 & 13.72 & 76.39 & 8.23 \\
& Random & 76.72 & 8.59 & 69.84 & 15.47 & 85.70 & 11.48 & 84.78 & 13.47 & 80.28 & 8.05 \\
& Major & 78.84 & 9.69 & \textbf{71.98} & 15.25 & \textbf{85.87} & 10.87 & 84.78 & \textcolor{gray}{\bf 12.36} & \textbf{80.31} & 6.98 \\
& Degree & \textbf{80.95} & 7.15 & 70.26 & 15.23 & 85.51 & 11.21 & \textbf{84.79} & 13.44 & 79.53 & \textcolor{gray}{\bf 6.90} \\
\bottomrule
    \end{tabular}
\end{table*}

\noindent
{\bf \large Experimental Settings.}
Before answering these questions, we first introduce the experimental settings. \\
\noindent
{\bf Datasets.}
Tab. \ref{table:dataset} presents statistics of five public node-labeled graph datasets. \textbf{Cora}, \textbf{Citeseer}, and \textbf{Pubmed} are famous citation graph data \citep{sen2008collective}. \textbf{AMZcobuy} comes from the photo segment of the Amazon co-purchase graph \cite{shchur2018pitfalls}. \textbf{Coauthor} is co-authorship graphs of computer science based on the Microsoft Academic Graph from the KDD Cup 2016 challenge \footnote{https://www.kdd.org/kdd-cup/view/kdd-cup-2016}.
For all datasets, the percentage of train, validation, and test partition comprise 10\%, 20\%, and 70\% of the nodes, respectively. Similar to GraphSS~\cite{zhuang2022defending}, only the train graphs contain manual-annotated labels, but the validation and test graphs don't. We simulate the manual-annotated labels by randomly replacing the ground-truth labels of 10\% (noise ratio $nr$) nodes with another label, chosen uniformly.

\noindent
{\bf Implementation of Topological Perturbations.}
We examine our model under three scenarios of topological perturbations as follows.
1) Random perturbations ({\bf rdmPert}): perturbators in OSNs intend to randomly connect with many other normal users for commercial purposes, e.g., new brand promotion. In the experiments, we limit the number of perturbators to 1\% of the validation/test nodes (a.k.a. victim nodes) and restrict the number of connections from each perturbator to 100. Note that rdmPert doesn't apply gradient-based attacks, such as FGSM \cite{goodfellow2014explaining}, PGD \cite{madry2017towards}, etc.
2) Information sparsity ({\bf infoSparse}): we sparse the 90\% links and 100\% features of the victim nodes ($L\&F$) on validation/test graphs.
3) Adversarial attacks ({\bf advAttack}): we execute node-level direct evasion non-targeted attacks~\cite{zugner2018adversarial} on both links and features ($L\&F$) of the victim nodes on sparse graphs, whereas the trained node classifier remains unchanged. To ensure the sparsity, we sparse the denser graphs (AMZcobuy and Coauthor) and select the victim nodes whose total degrees are within (0, 10) for attacks. The intensity of perturbation $n_{pert}$ is set as 2 for all datasets. The ratio of $n_{pert}$ between applying on links and applying on features is 1: 10.

\begin{figure*}[t]  
  \hfill
  \begin{subfigure}{0.195\textwidth}
  \centering 
    \includegraphics[width=\linewidth]{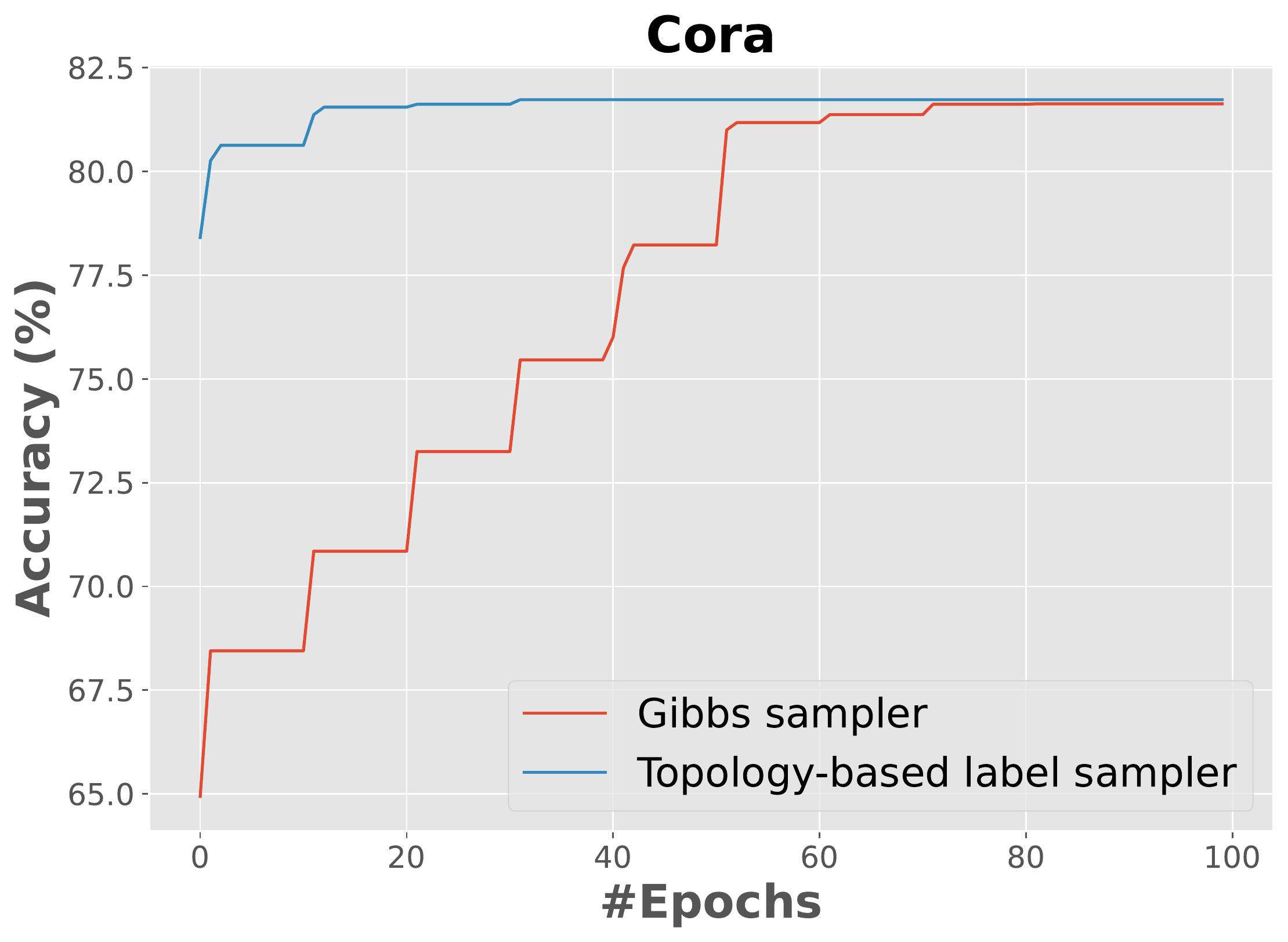}
  \end{subfigure}%
  \hfill
  \begin{subfigure}{0.195\textwidth}
  \centering 
    \includegraphics[width=\linewidth]{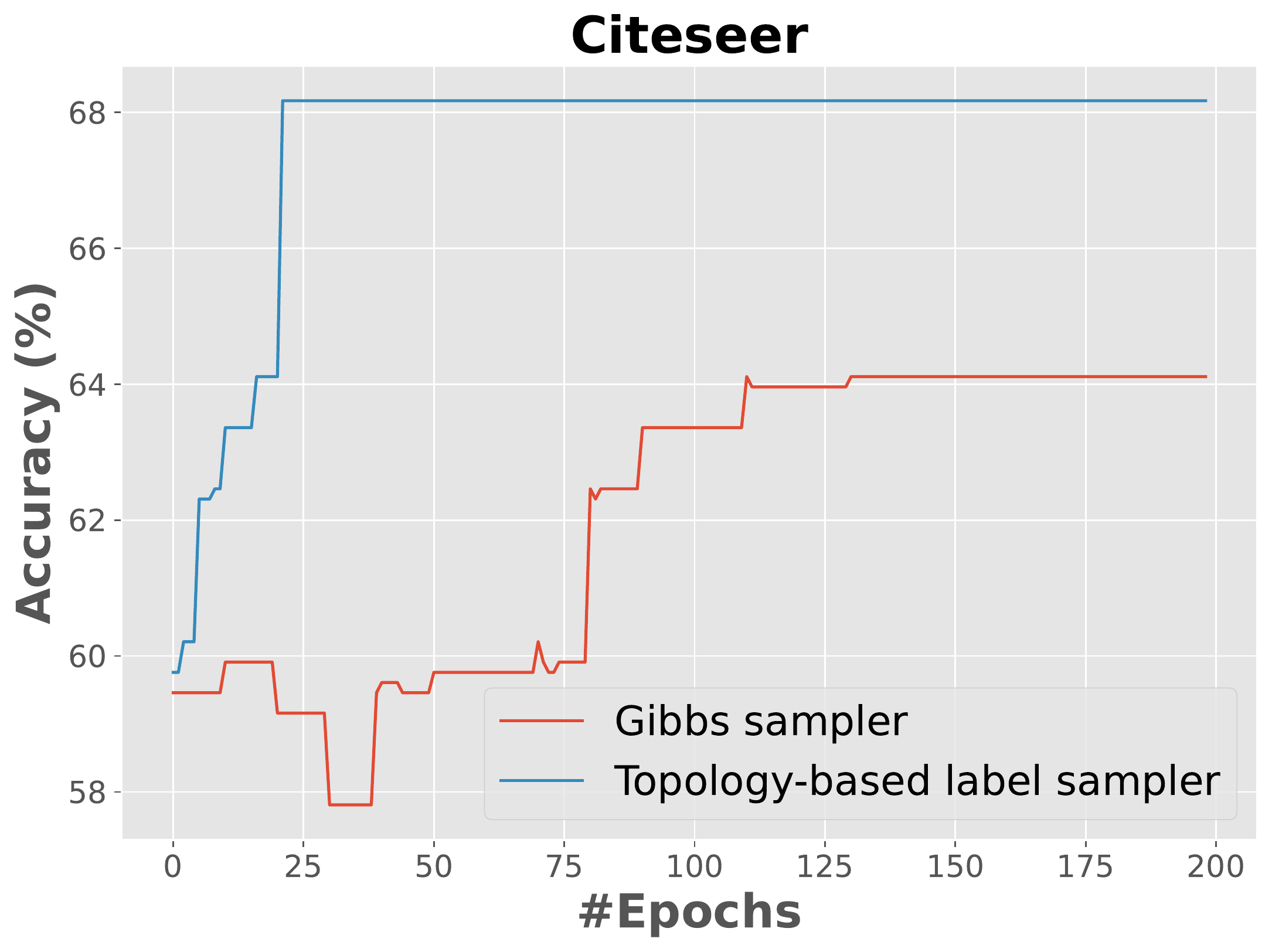}
  \end{subfigure}
  \begin{subfigure}{0.195\textwidth}
  \centering 
    \includegraphics[width=\linewidth]{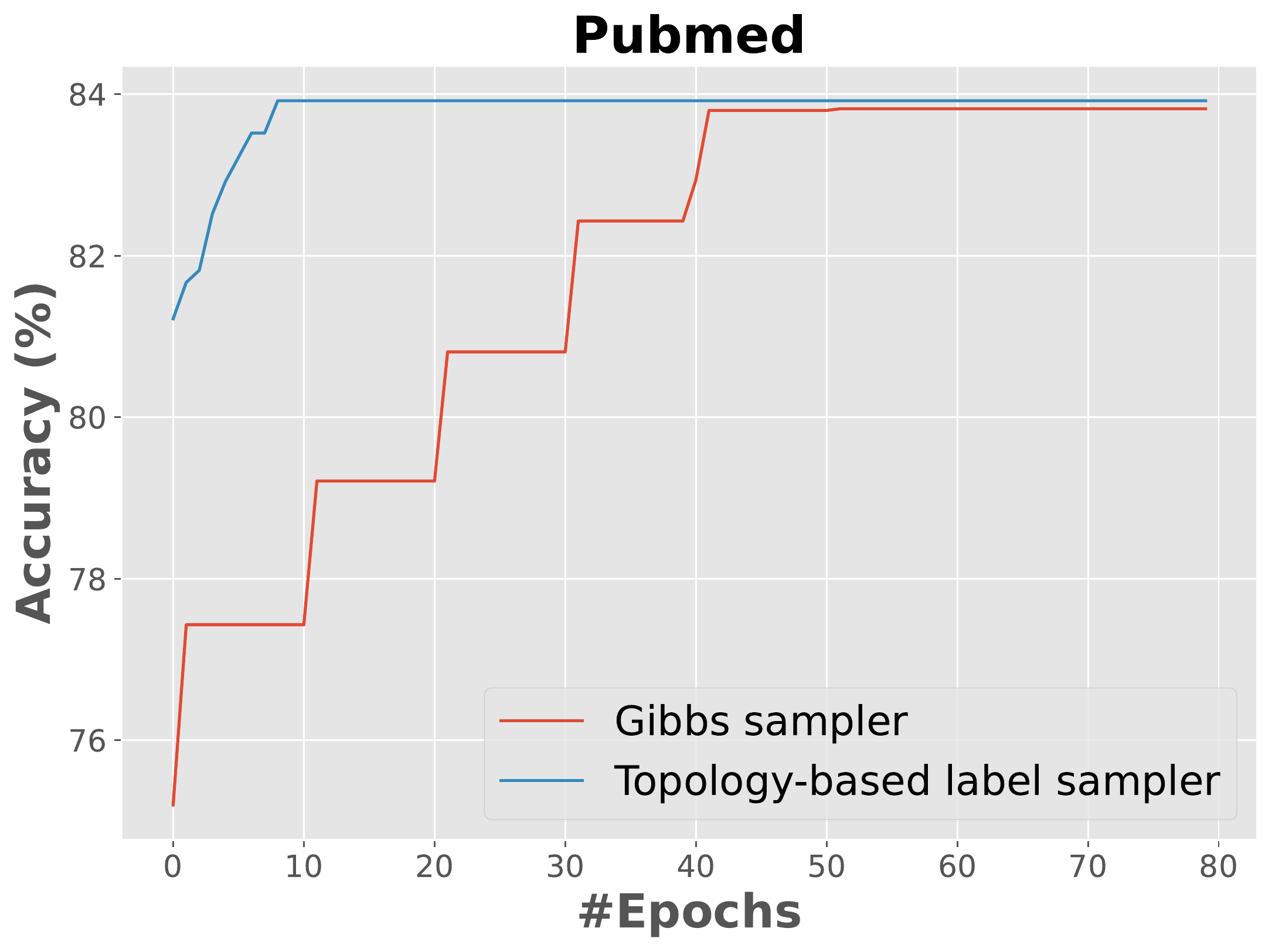}
  \end{subfigure}
  \begin{subfigure}{0.195\textwidth}
  \centering 
    \includegraphics[width=\linewidth]{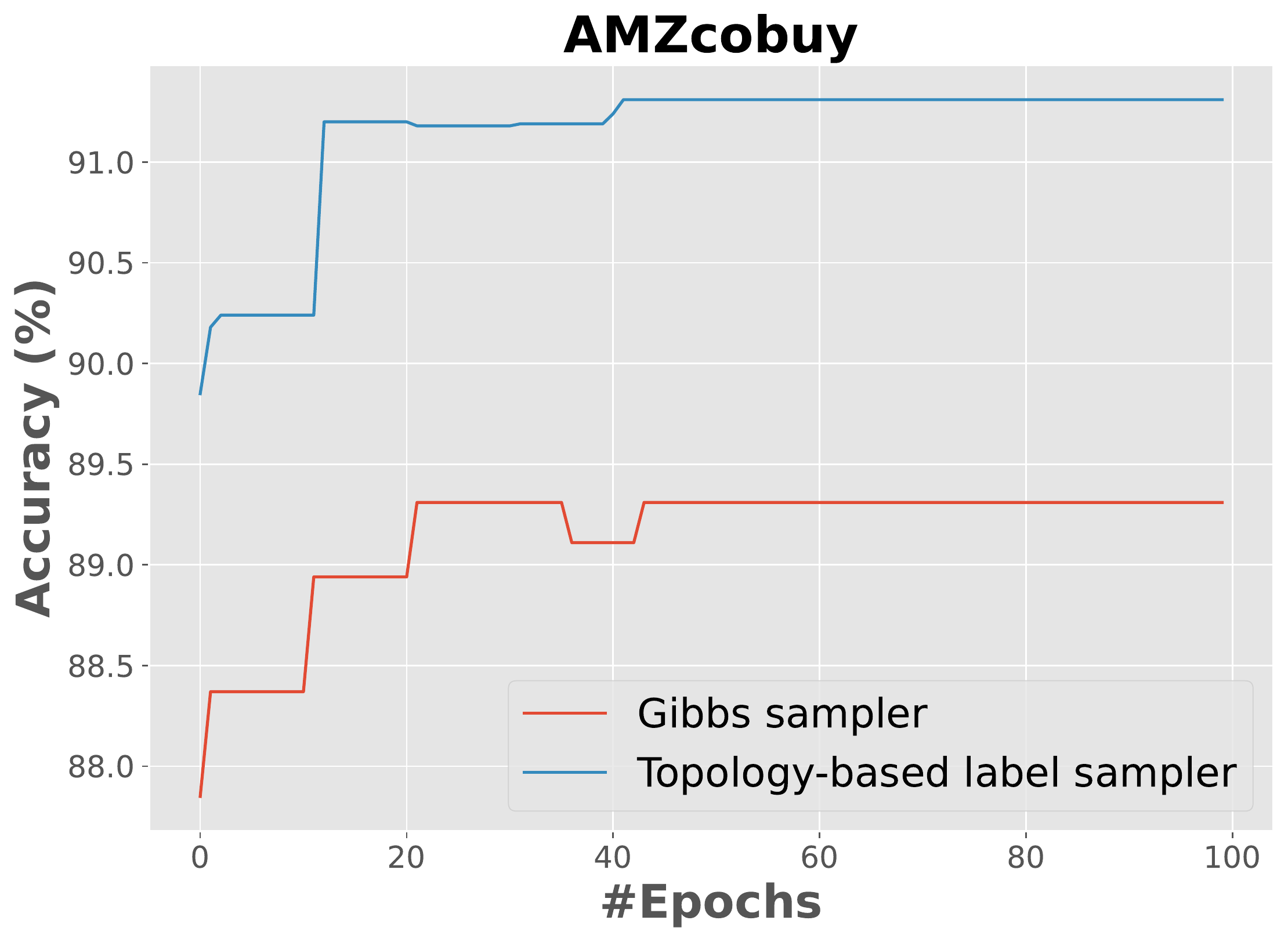}
  \end{subfigure}
  \begin{subfigure}{0.195\textwidth}
  \centering 
    \includegraphics[width=\linewidth]{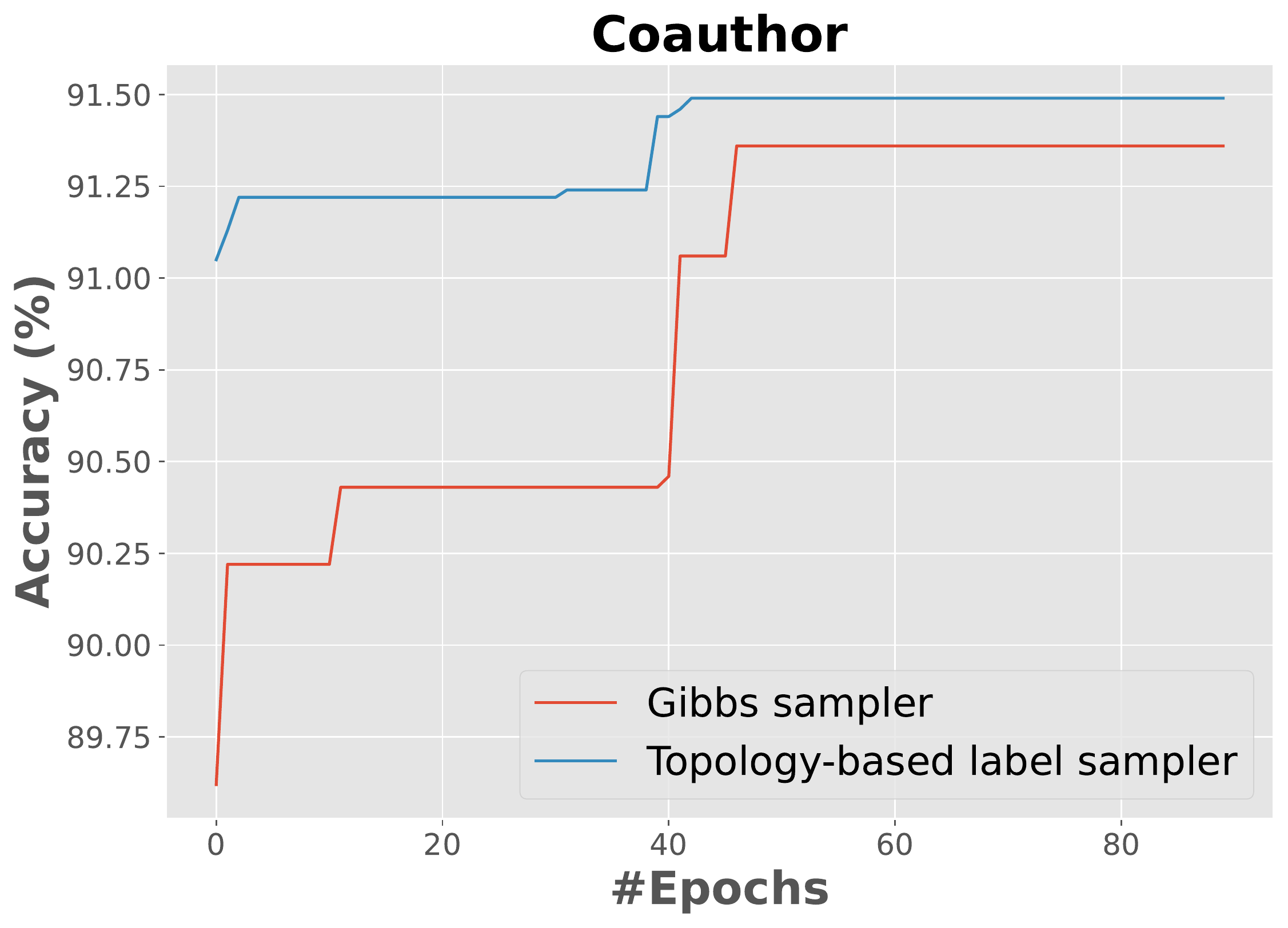}
  \end{subfigure}
  \hfill
  \begin{subfigure}{0.195\textwidth}
  \centering 
    \includegraphics[width=\linewidth]{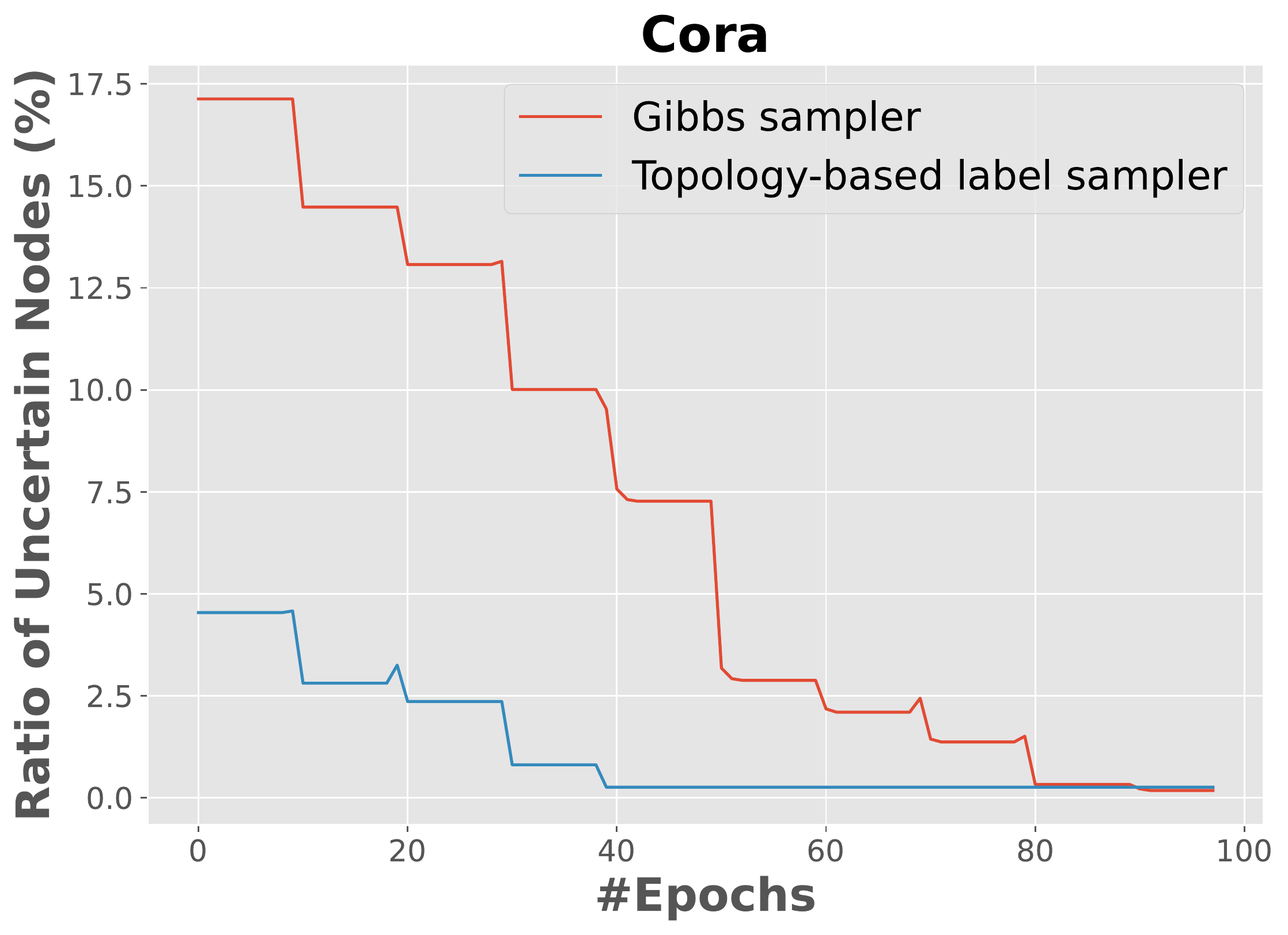}
  \end{subfigure}%
  \hfill
  \begin{subfigure}{0.195\textwidth}
  \centering 
    \includegraphics[width=\linewidth]{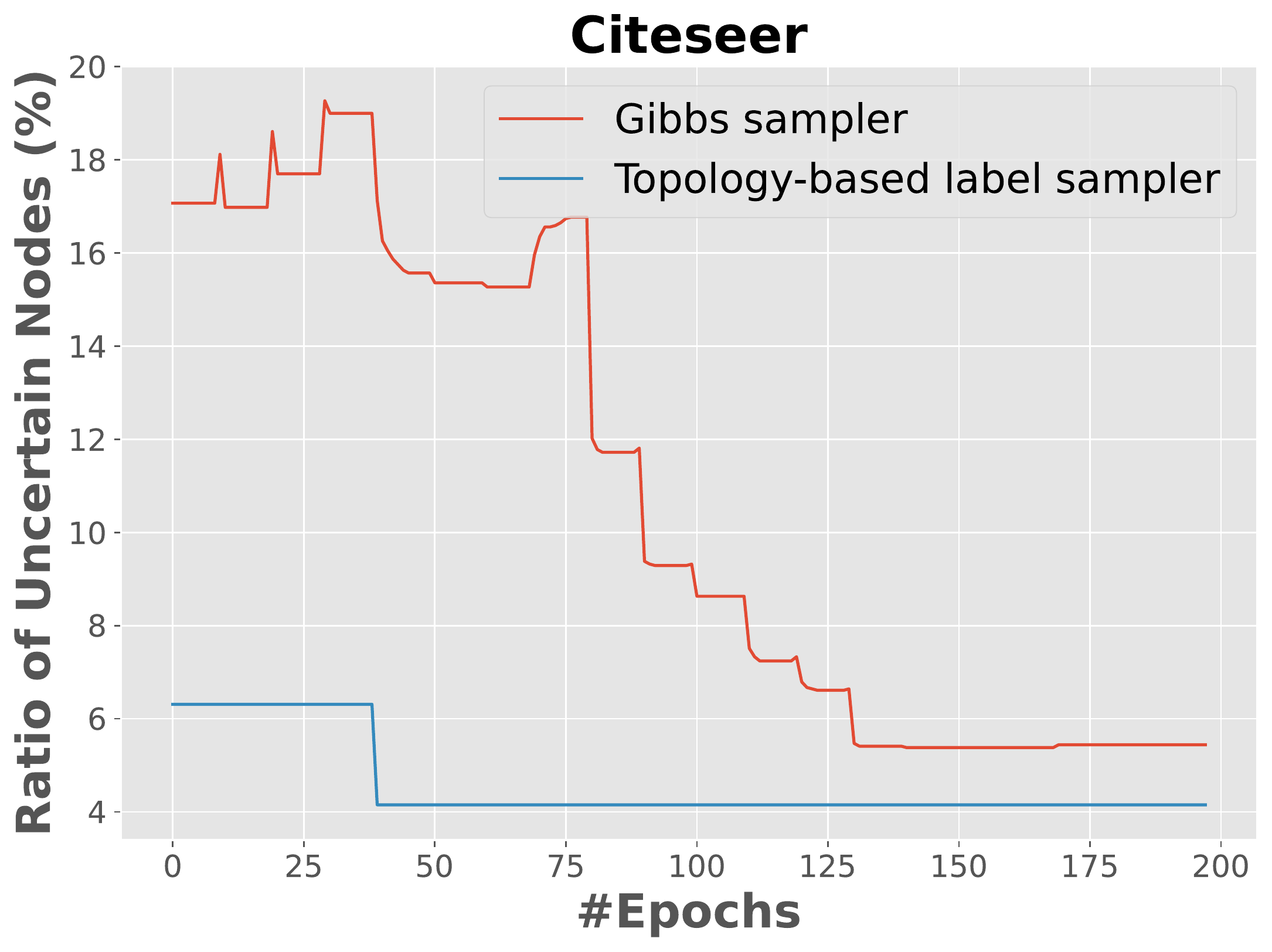}
  \end{subfigure}
  \begin{subfigure}{0.195\textwidth}
  \centering 
    \includegraphics[width=\linewidth]{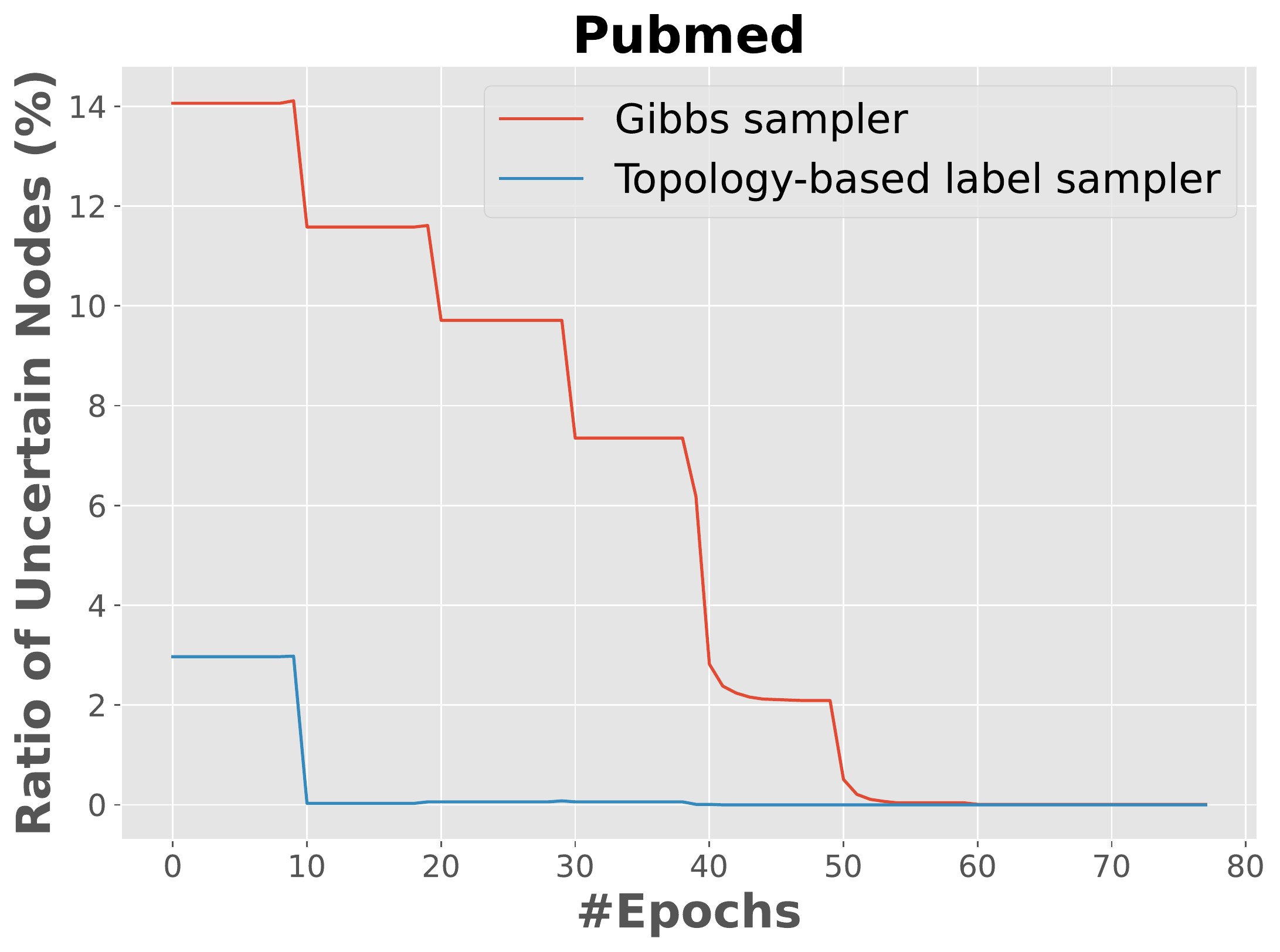}
  \end{subfigure}
  \begin{subfigure}{0.195\textwidth}
  \centering 
    \includegraphics[width=\linewidth]{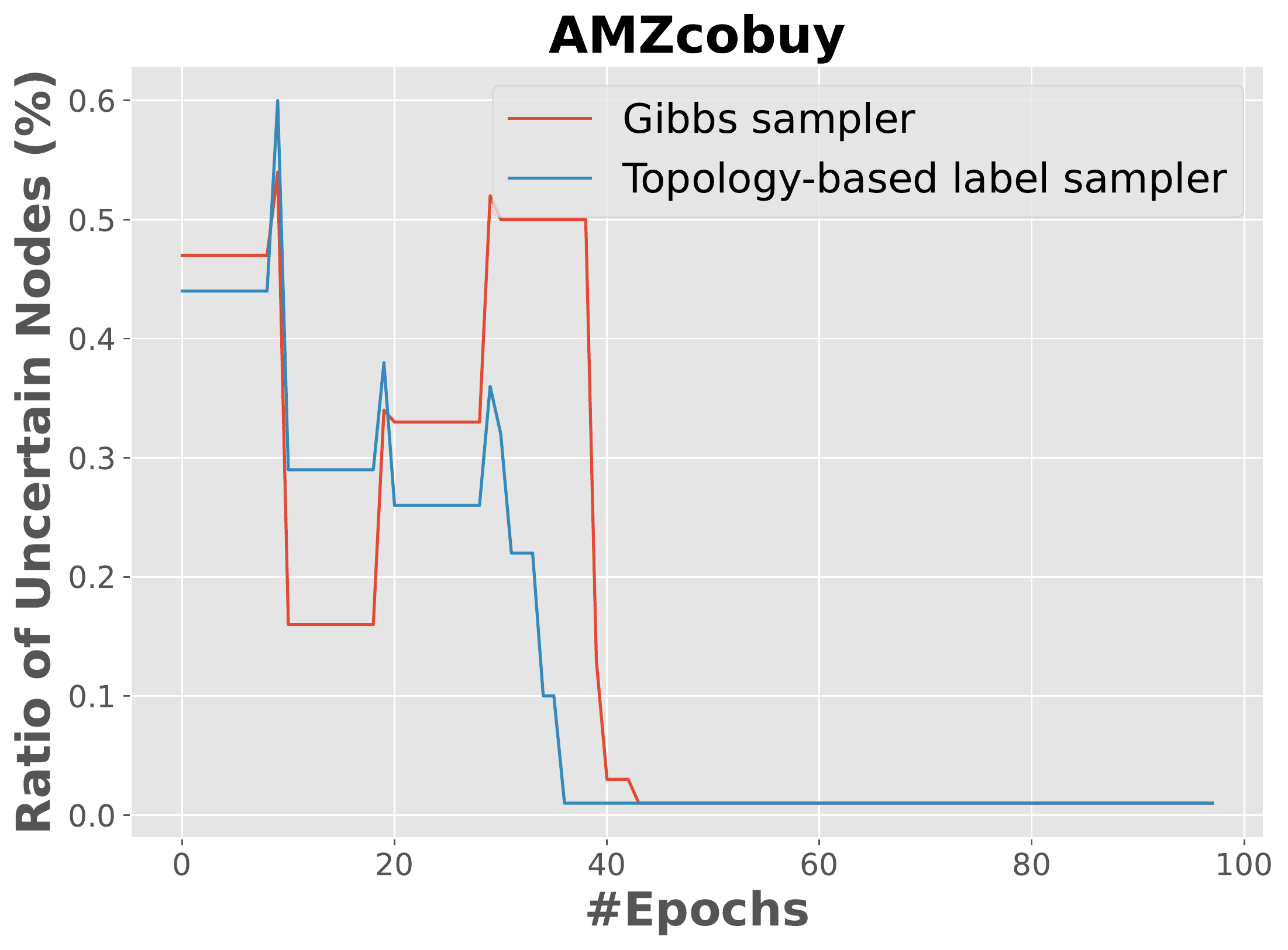}
  \end{subfigure}
  \begin{subfigure}{0.195\textwidth}
  \centering 
    \includegraphics[width=\linewidth]{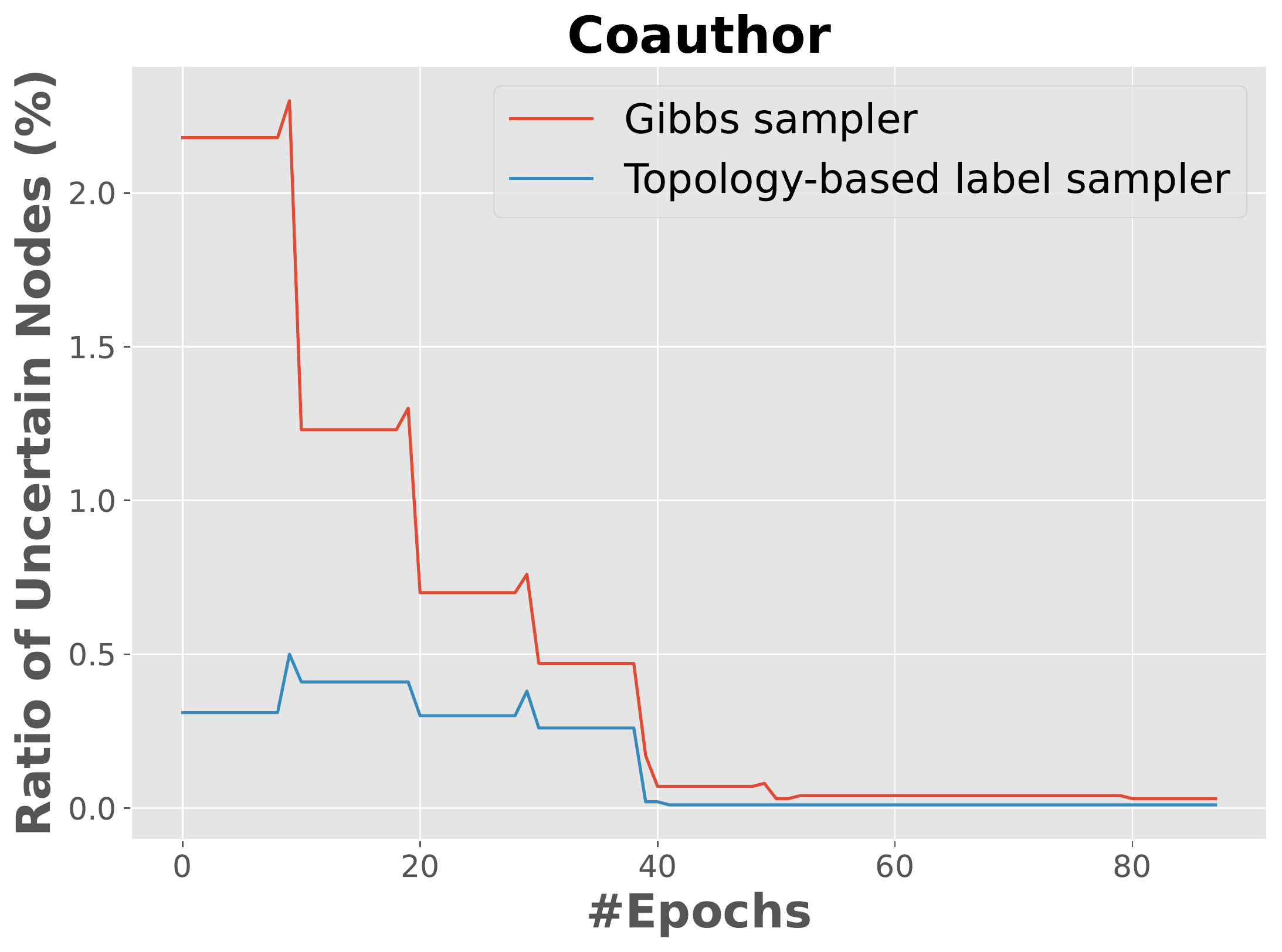}
  \end{subfigure}
\caption{Empirical analysis of convergence between Gibbs sampler and Topology-based label sampler (Major).}
\label{fig:fig_convergence}
\end{figure*}

\begin{figure}[h] 
  \centering
  \includegraphics[width=\linewidth]{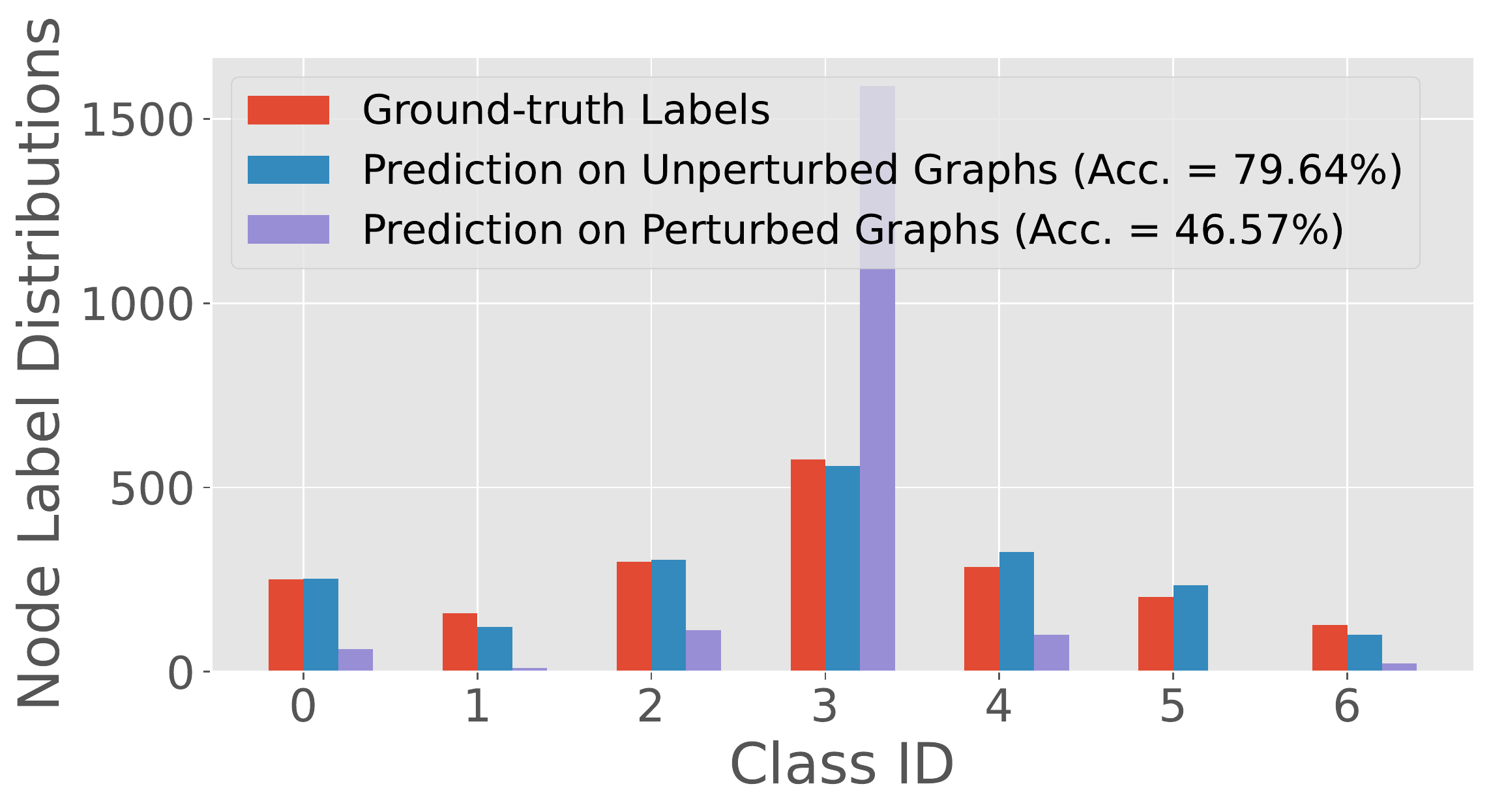}
  \caption{Node label distributions of Cora (test nodes).}
\label{fig:fig_dist}
\end{figure}

\noindent
{\bf Competing Methods.}
We first compare our model with seven popular competing methods which develop robust GNN-based node classifiers using topological information. Also, we consider MC Dropout as one of the competing methods for measuring uncertainty purposes.
{\bf GNN-Jaccard}~\cite{wu2019adversarial} preprocesses the graph by eliminating suspicious connections, whose Jaccard similarity of node’s features is smaller than a given threshold.
{\bf GNN-SVD}~\cite{entezari2020all} proposes another preprocessing approach with low-rank approximation on the perturbed graph to mitigate the negative effects from high-rank attacks, such as Nettack~\citep{zugner2018adversarial}.
{\bf DropEdge}~\cite{rong2019dropedge} randomly removes a number of edges from the input graph in each training epoch.
{\bf GRAND}~\cite{feng2020graph} proposes random propagation and consistency regularization strategies to address the issues of over-smoothing and non-robustness of GCNs.
{\bf RGCN}~\cite{zhu2019robust} adopts Gaussian distributions as the hidden representations of nodes to mitigate the negative effects of adversarial attacks.
{\bf ProGNN}~\cite{jin2020graph} jointly learns the structural graph properties and iteratively reconstructs the clean graph to reduce the effects of adversarial structure.
{\bf GDC}~\cite{hasanzadeh2020bayesian} proposes an adaptive connection sampling method using stochastic regularization for training GNNs. This method can learn with uncertainty on graphs.
{\bf MC Dropout}~\cite{gal2016dropout} develops a dropout framework that approximates Bayesian inference in deep Gaussian processes.

\noindent
{\bf Evaluation Metrics.}
We use both accuracy (Acc.) and average normalized entropy (Ent.) to measure the robustness of a node classifier. More specifically, Acc. evaluates the node classification performance (the higher the better), whereas Ent. represents the uncertainty of nodes' categorical distributions (the lower the better).

\noindent
{\bf \large Node Classification Benefits from Our Model against Topological Perturbations.}
The performance of a node classifier may degrade on perturbed graphs. We present the comparison of node label distributions of test nodes among the ground-truth labels, the prediction on unperturbed graphs, and the prediction on perturbed graphs (under rdmPert) in Fig. \ref{fig:fig_dist} as an example. This comparison indicates that node label distributions may significantly change on perturbed graphs, causing performance degradation.
In this study, our goal is to improve the robustness of the node classifier on perturbed graphs. In other words, we aim to increase the classification accuracy (Acc.) while maintaining the uncertainty (Ent.) at a low level.

To answer the first question, we examine whether the node classifier can benefit from our model under three scenarios of topological perturbations, random perturbations (rdmPert), information sparsity (infoSparse), and adversarial attacks (advAttack). We compare the performance between the baseline model, GraphSS~\cite{zhuang2022defending}, and the variants of our model architectures. The denotations of related methods are mentioned in the caption of Tab. \ref{table:exp1}.
The results affirmatively verify that our model can outperform the baseline model and successfully achieve our goal in most cases. 
Besides, we have several observations as follows.
At first, topological samplers could further boost accuracy under both rdmPert and advAttack but may fail under infoSparse because most links and features are sparsified under this scenario.
Moreover, the majority sampler attains higher accuracy than the degree-weighted sampler on sparse graphs (Cora, Citeseer, and Pubmed) in most cases. Such a situation may be largely reversed on denser graphs (AMZcobuy and Coauther) because the degree-weighted sampler can further take the advantage of degree information from neighbors.
Furthermore, the uncertainty may be lower under some scenarios, e.g., Ent. is 1.43\% under advAttack on Cora. However, such kind of low uncertainty reveals that the node classifier may suffer severe perturbations, i.e., many nodes are certainly assigned to incorrect classes. Thus, lower uncertainty couldn't fully indicate robust performance.
Last but not least, our model couldn't significantly decrease the uncertainty under rdmPert on Citeseer.

\begin{figure}[h] 
  \centering
  \includegraphics[width=\linewidth]{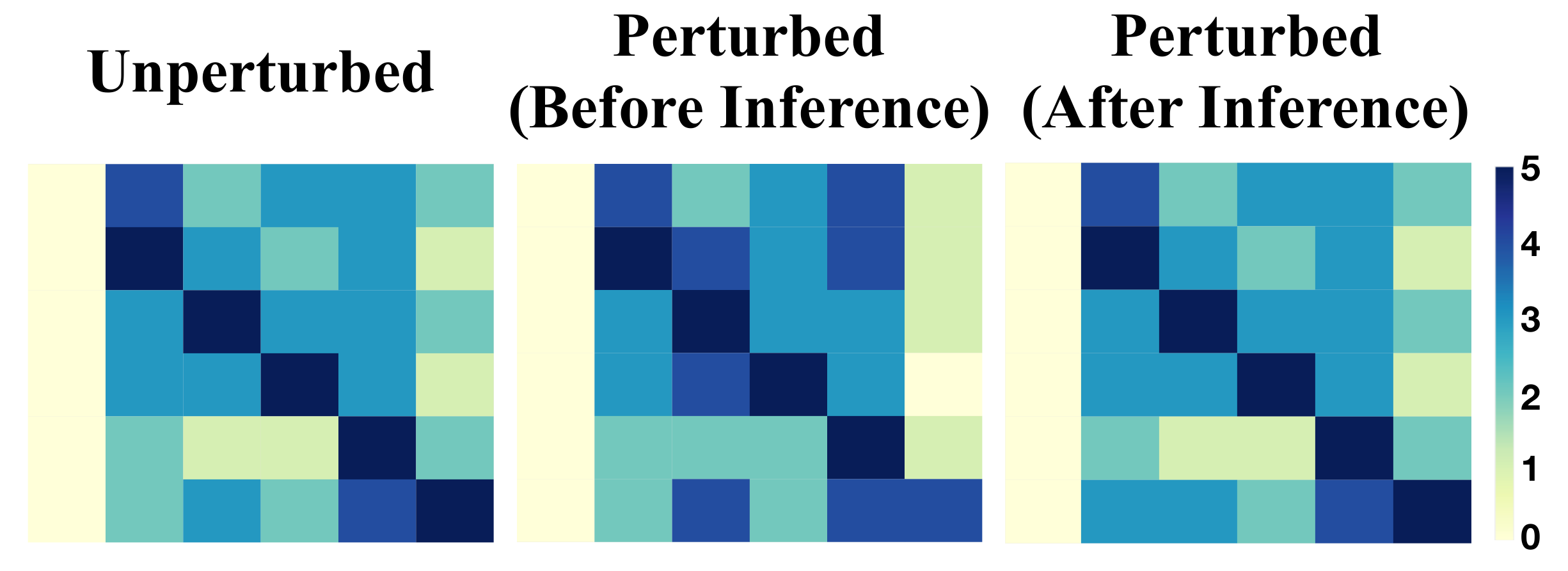}
  \caption{Visualization of node label distributions on Citeseer via log-scale fine-grained confusion matrices.}
\label{fig:fig_cm}
\end{figure}

\begin{table*} 
\small
\centering
\setlength{\tabcolsep}{4.5pt}
\caption{Comparison between competing methods and our model under the random perturbations scenario. Acc. (\%) denotes classification accuracy. Ent. (\%) denotes average normalized entropy. Time (s) denotes total runtime.}
\label{table:exp4}
\begin{tabular}{c|ccc|ccc|ccc|ccc|ccc} 
\toprule 
\multirow{2}{*}{\parbox{1.3cm}{\centering {\bf Methods}}} & \multicolumn{3}{c|}{{\bf Cora}} & \multicolumn{3}{c|}{{\bf Citeseer}} & \multicolumn{3}{c|}{{\bf Pubmed}} & \multicolumn{3}{c|}{{\bf AMZcobuy}} & \multicolumn{3}{c}{{\bf Coauthor}} \\
\cline{2-16}
& {\bf Acc.} & {\bf Ent.} & {\bf Time} & {\bf Acc.} & {\bf Ent.} & {\bf Time} & {\bf Acc.} & {\bf Ent.} & {\bf Time} & {\bf Acc.} & {\bf Ent.} & {\bf Time} & {\bf Acc.} & {\bf Ent.} & {\bf Time} \\
\midrule
{\bf GNN-Jaccard}~\cite{wu2019adversarial} & 66.32 & 93.24 & 2.83 & 56.65 & 95.47 & 2.34 & 60.68 & 81.77 & 3.71 & 53.12 & 96.17 & 5.83 & 88.47 & 96.88 & 7.44 \\
{\bf GNN-SVD}~\cite{entezari2020all} & 50.53 & 93.01 & 3.05 & 31.76 & 95.20 & 3.48 & 74.22 & 88.39 & 12.96 & 70.02 & 93.52 & 5.01 & 74.22 & 97.21 & 11.51 \\
{\bf DropEdge}~\cite{rong2019dropedge} & 67.88 & 95.28 & \textcolor{gray}{\bf 1.89} & 46.78 & 96.44 & \textcolor{gray}{\bf 1.46} & 77.34 & 76.66 & 2.56 & 63.42 & 96.25 & 2.61 & 71.42 & 97.48 & \textcolor{gray}{\bf 2.90} \\
{\bf GRAND}~\cite{feng2020graph} & 52.34 & 94.99 & 8.72 & 35.02 & 95.35 & 5.69 & 49.75 & 89.89 & 12.15 & 40.22 & 96.23 & 27.14 & 55.56 & 97.73 & 154.17 \\
{\bf RGCN}~\cite{zhu2019robust} & 62.63 & 93.72 & 5.04 & 62.23 & 96.39 & 5.17 & 83.20 & 89.91 & 27.71 & 77.87 & 97.36 & 30.85 & 89.33 & 98.39 & 179.65 \\
{\bf ProGNN}~\cite{jin2020graph} & 52.63 & 92.09 & 174.75 & 36.18 & 96.32 & 294.93 & 50.11 & 88.35 & 2135.67 & 45.28 & 98.63 & 1914.42 & 57.39 & 96.55 & 2366.04 \\
{\bf GDC}~\cite{hasanzadeh2020bayesian} & 71.18 & 85.77 & 12.30 & 43.15 & 93.73 & 17.52 & 48.95 & 32.94 & 258.34 & 45.58 & 98.18 & 80.84 & 62.65 & 94.02 & 205.22 \\
{\bf MC Dropout}~\cite{gal2016dropout} & 80.53 & 40.25 & 2.41 & 64.81 & 89.80 & 2.74 & 79.51 & 72.73 & \textcolor{gray}{\bf 1.99} & 90.68 & 39.21 & \textcolor{gray}{\bf 1.81} & 88.40 & 43.30 & 5.79 \\
{\bf Ours} & {\bf 84.19} & \textcolor{gray}{\bf 23.54} & 29.35 & {\bf 67.84} & \textcolor{gray}{\bf 62.50} & 70.73 & {\bf 83.57} & \textcolor{gray}{\bf 32.57} & 158.25 & {\bf 91.92} & \textcolor{gray}{\bf 12.55} & 75.28 & {\bf 90.60} & \textcolor{gray}{\bf 7.08} & 184.15 \\
\bottomrule
    \end{tabular}
\end{table*}

\begin{figure*}[t]  
  \hfill
  \begin{subfigure}{0.195\textwidth}
  \centering 
    \includegraphics[width=\linewidth]{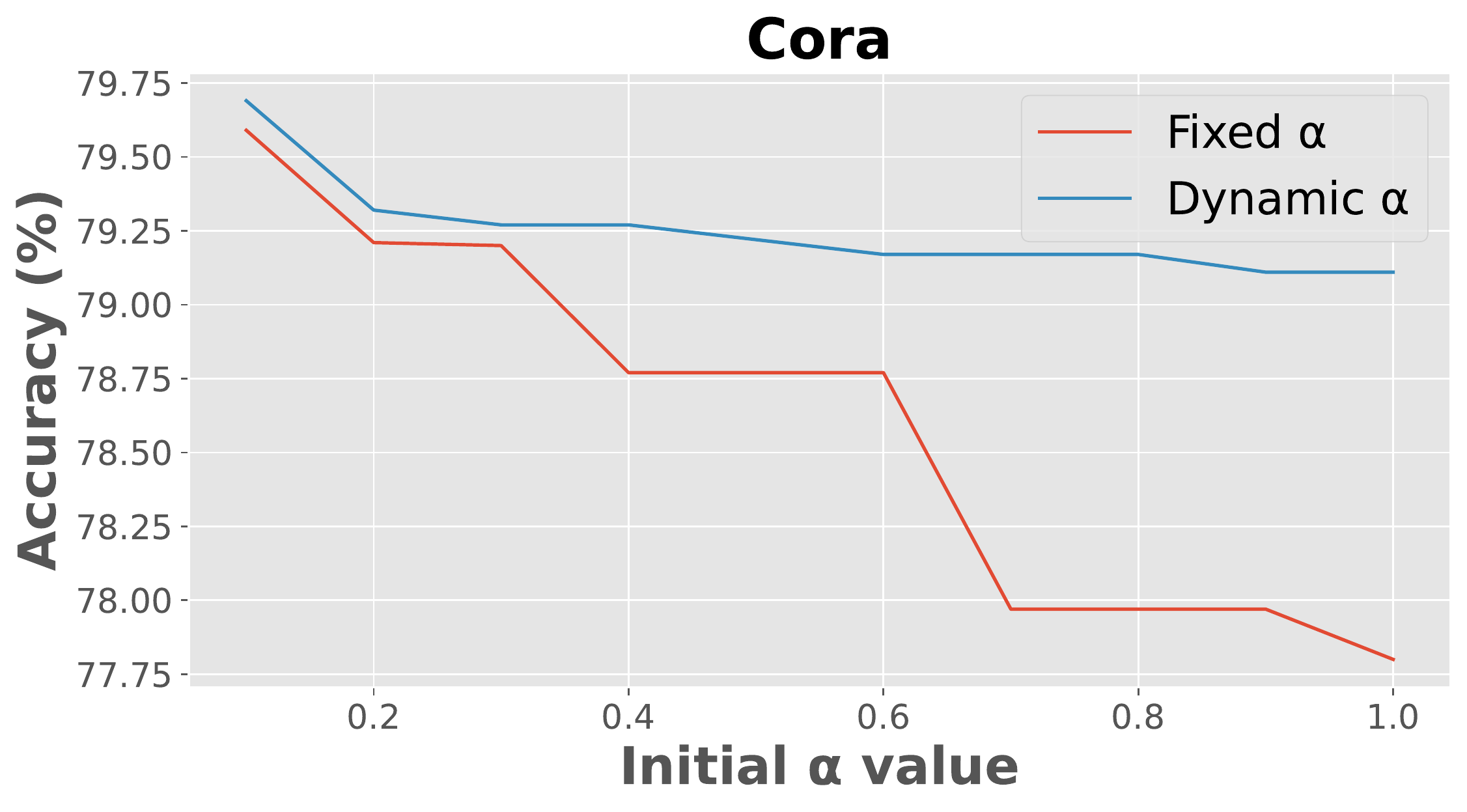}
  \end{subfigure}
  \hfill
  \begin{subfigure}{0.195\textwidth}
  \centering 
    \includegraphics[width=\linewidth]{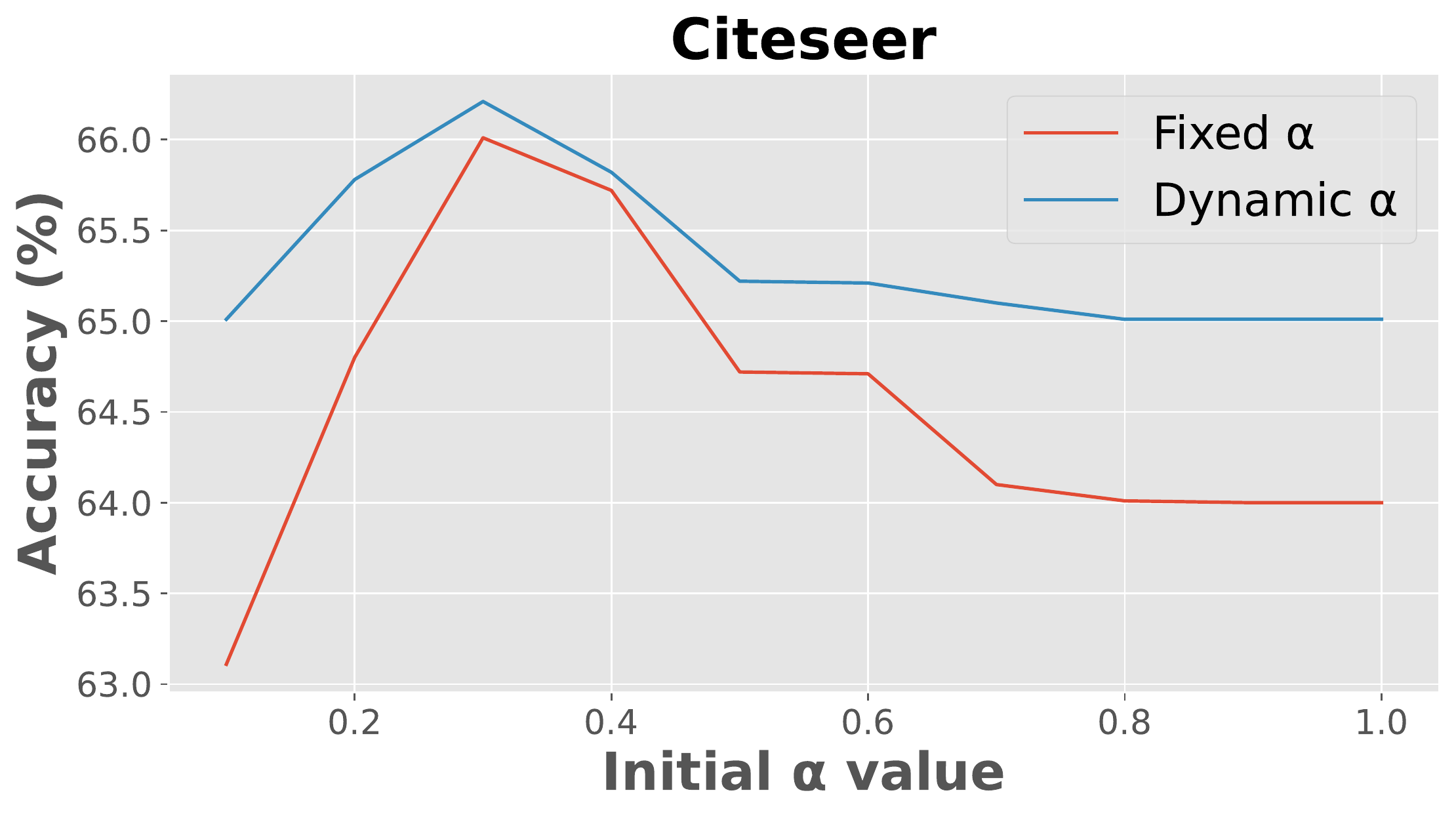}
  \end{subfigure}
  \begin{subfigure}{0.195\textwidth}
  \centering 
    \includegraphics[width=\linewidth]{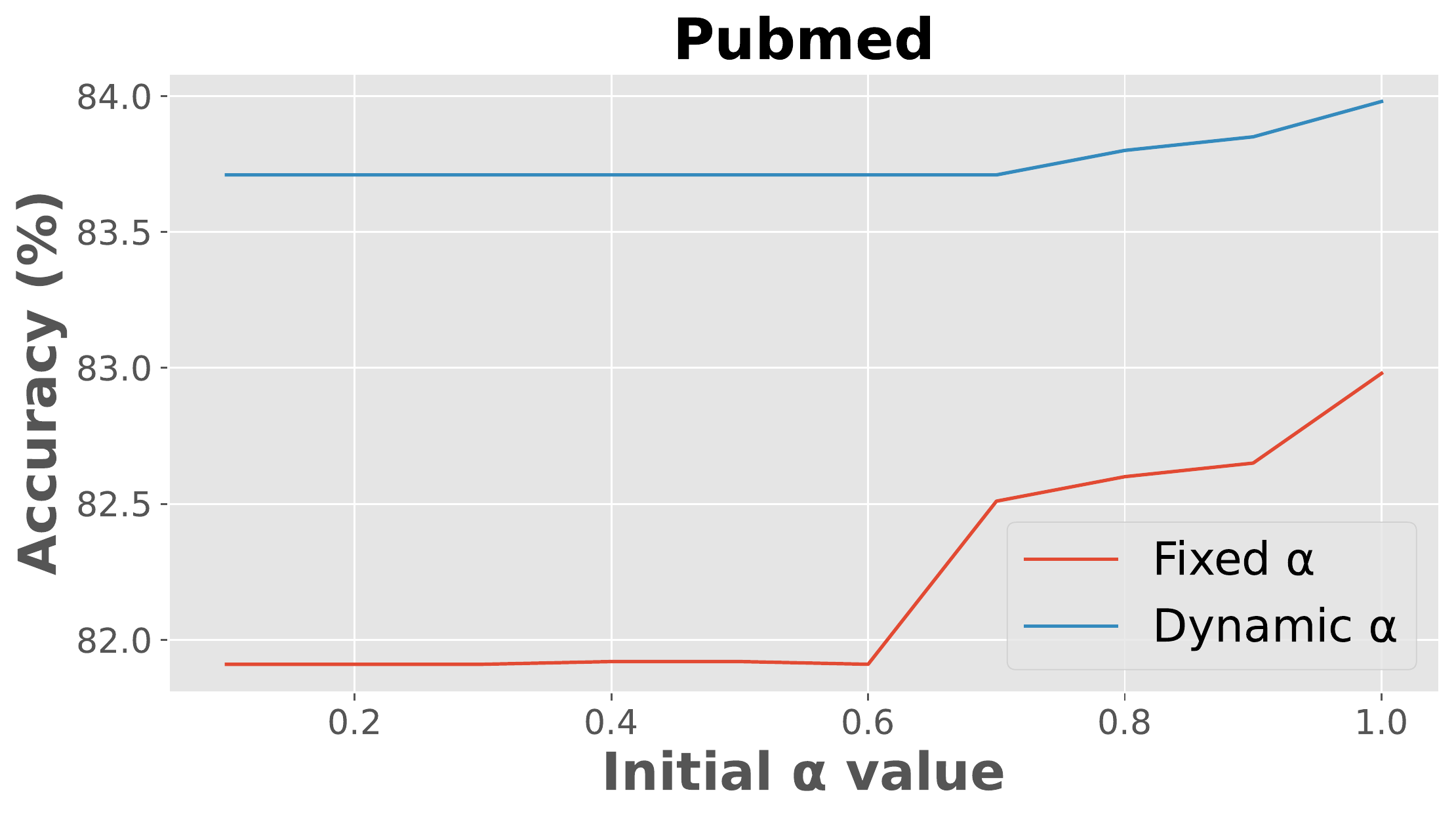}
  \end{subfigure}
  \begin{subfigure}{0.195\textwidth}
  \centering 
    \includegraphics[width=\linewidth]{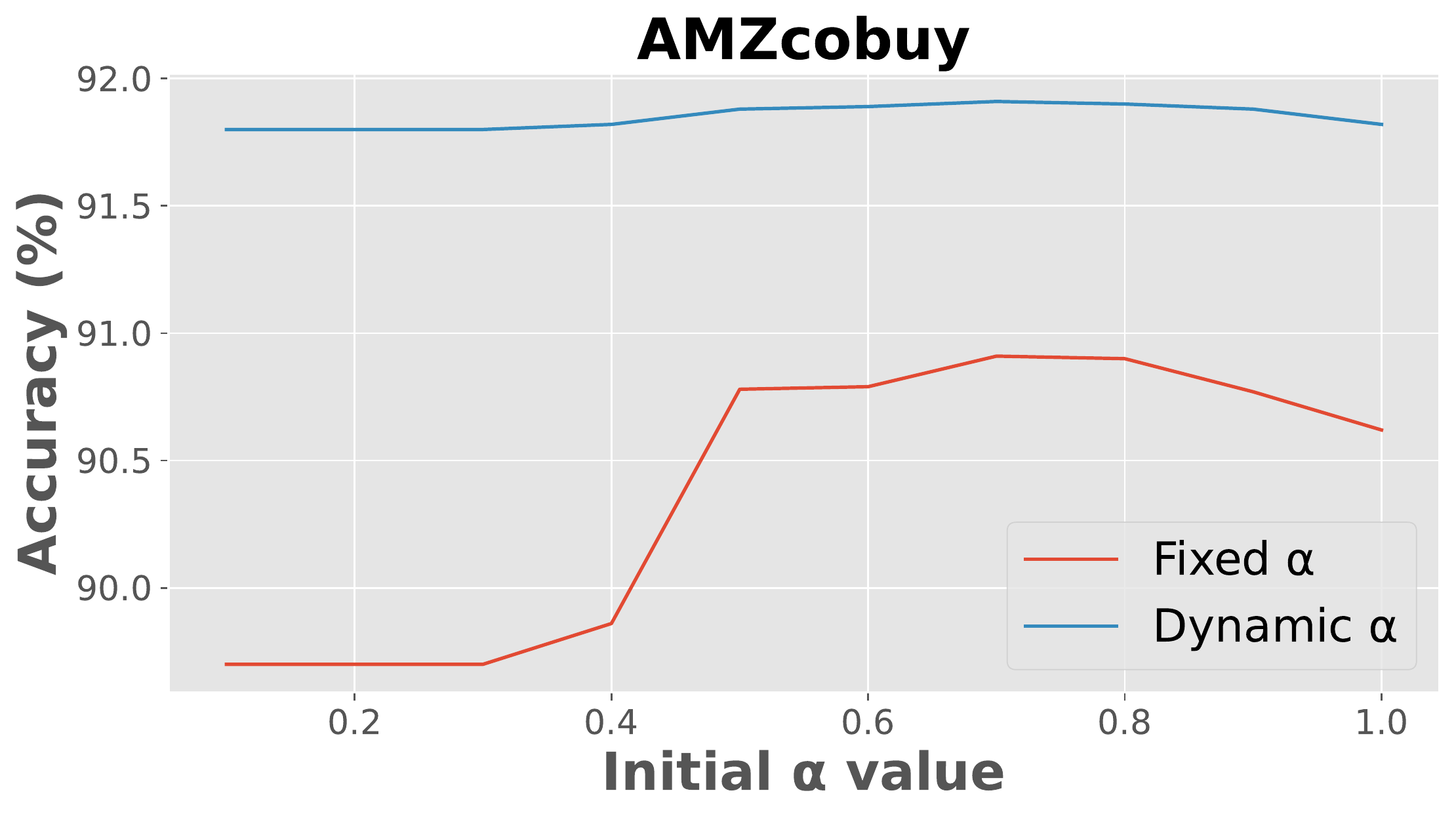}
  \end{subfigure}
  \begin{subfigure}{0.195\textwidth}
  \centering 
    \includegraphics[width=\linewidth]{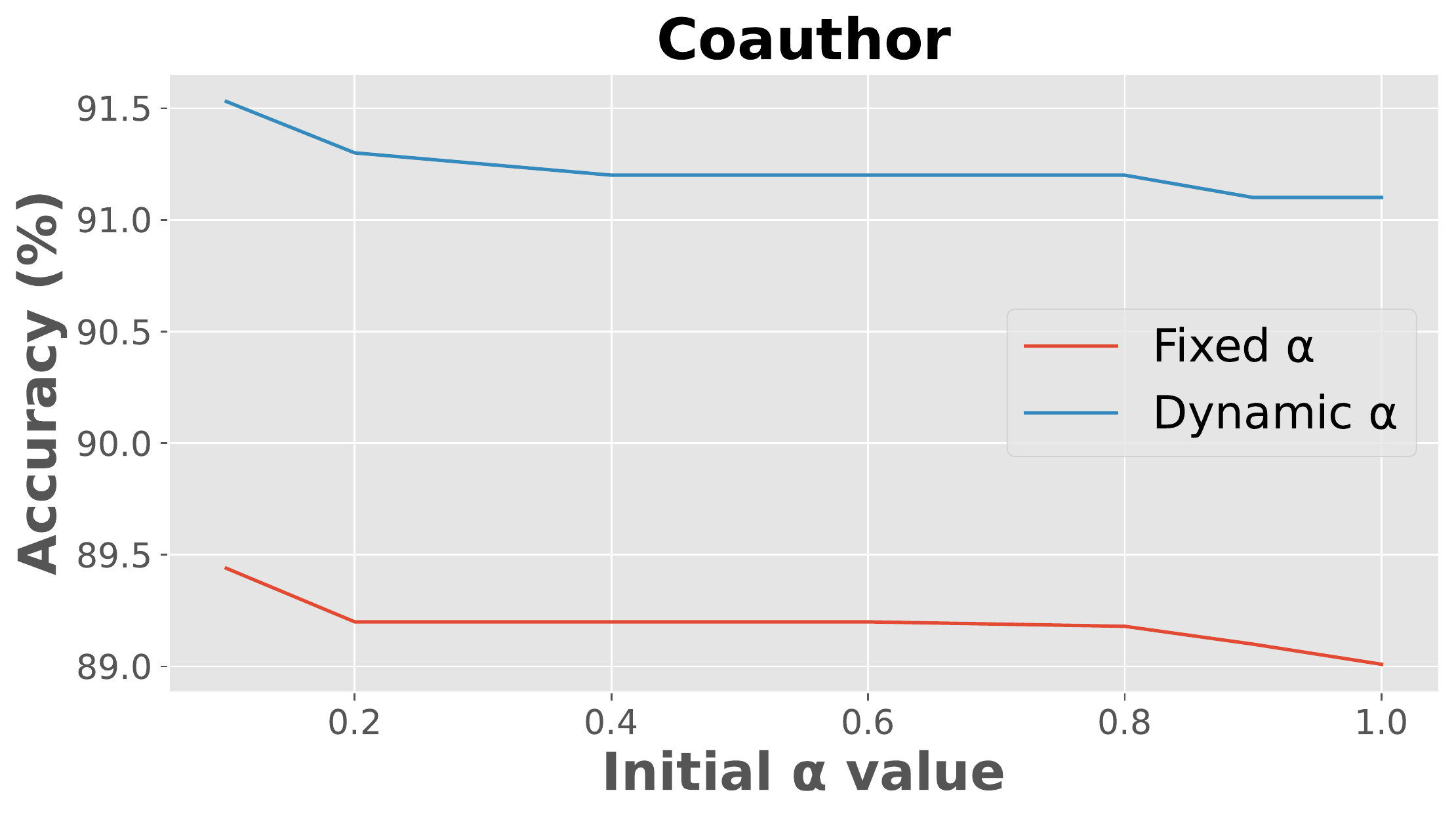}
  \end{subfigure}
\caption{Analysis of the initial $\alpha$ value for the dynamic $\alpha$ vector.}
\label{fig:fig_hp}
\end{figure*}

We further investigate this unusual case by visualizing the node label distributions via confusion matrices in Fig. \ref{fig:fig_cm}. The left matrix presents the predicted node label distributions on unperturbed graphs, whereas the middle and right matrices present the predicted and inferred distributions on perturbed graphs, respectively.
The visualization first shows that our model can recover the distribution as close as that on unperturbed graphs when the node label distribution is perturbed.
These confusion matrices also indicate that the node classification on the unperturbed graphs already misses one class, leading to the inaccurate prior distribution. The label inference may probably fail to decrease the uncertainty due to this reason.
Besides, this case reveals two limitations of our model. First, the inference highly depends on an accurate prior distribution. Second, our model couldn't recover the missing class, which implies that our model currently cannot handle the open-set classification.

\noindent
{\bf \large Empirical Analysis of Convergence.}
To answer the second question, we conduct an empirical analysis of convergence to examine two proposed conjectures. Fig. \ref{fig:fig_convergence} displays the empirical comparison between using the Gibbs sampler and using the topology-based label sampler (Major) during the label inference under rdmPert. The first row presents the curves of validation accuracy (\%). The second row displays the curves of the ratio of uncertain nodes (\%) on inferred labels. According to the results, the label inference using both sampling methods can eventually converge across five datasets. The convergence verifies that the first conjecture holds true. Also, the results show that the label inference using the topology-based label sampler reaches the convergence with fewer iterations of transition compared to using the Gibbs sampler. In other words, using the topology-based label sampler can help decrease the ratio of uncertain nodes faster, and further reduces the fluctuations of the curve. The comparison verifies the second conjecture.

\noindent
{\bf \large Comparison with Competing Methods.}
Tab. \ref{table:exp4} presents the comparison of performance between competing methods and our model under rdmPert. For our model, we present the average values of three topological samplers. The results verify that our model is superior to competing methods across five datasets. Most competing methods fail to decrease the uncertainty (Ent.) but MC Dropout obtains robust performance among them. Our model can outperform MC Dropout, achieving higher classification accuracy with lower uncertainty.
We notice that DropEdge and RGCN perform higher accuracy on larger graphs. For DropEdge, randomly dropping a certain number of edges is equivalent to data augmentation, which will be performed stronger when the size of the input graph is getting larger. RGCN adopts Gaussian distributions in hidden layers to reduce the negative impacts that come from the shift of node label distribution. The Gaussian-based hidden matrix has a higher capability to mitigate the negative impacts on larger graphs.

From the perspective of total runtime (in second), we observe that topological denoising methods (GCN-Jaccard, GCN-SVD, and DropEdge) and MC dropout consume much less time than other methods. These are reasonable because topological denoising methods mainly process the input graph structure, and MC dropout can speed up the training by dropping some hidden units. On the contrary, ProGNN consumes much more time as the size of graphs increases. One of the reasons is that ProGNN iteratively reconstructs the graph by preserving low rank and sparsity properties. Such reconstructions will take a much longer time on larger graphs.

\noindent
{\bf \large Analysis of Parameters.}
In this study, we maintain the same values of both warm-up steps $WS$, 40, and retraining epochs $Retrain$, 60, as that in GraphSS~\cite{zhuang2022defending}. To avoid over-fitting, we retrain the node classifier every 10 epochs during the inference. In the meanwhile, we analyze how the initial $\alpha$ value affects accuracy, and conduct this analysis on the validation set. We fix the number of transition states $T$ as [100, 200, 80, 100, 90] for five datasets, respectively.
Fig. \ref{fig:fig_hp} presents two curves of accuracy along different initial $\alpha$ values. The blue curve denotes the case that we dynamically update the $\alpha$ vector, whereas the red curve denotes the case with fixed $\alpha$ values.
The results affirmatively answer the fourth question that adopting asymmetric Dirichlet distributions as a prior can contribute to the label inference using Bayesian label transition. Besides, the comparison between two curves verifies that the dynamic $\alpha$ mechanism can further boost the accuracy in most cases. Based on this analysis, we select the initial $\alpha$ values as [0.1, 0.3, 1.0, 0.7, 0.1] for five datasets, respectively.

\noindent
{\bf \large Limitation and Future Directions.}
1) Using the topology-based sampler may fail to boost the classification accuracy on extreme sparse graphs where most links and features are sparsified or missing.
2) Our label inference model highly depends on an accurate prior distribution. In other words, our model couldn't handle the poisoning attacks. In the future, integrating a denoising approach on the prior distribution would be a possible solution to mitigate this issue.
3) Our model couldn't recover the missing class from the prior distribution. This limitation implies that our model currently cannot handle the open-set classification. In the future, we could jointly utilize a distance-based label transition matrix to detect potential unseen classes during the inference.

%% file: 2rewk.tex
\section{Related Work and Impact}
\label{sec:rewk}

In this section, we further discuss how our work may contribute to other research fields as follows:
{\bf 1) Noisy label learning.} Learning GNNs with noisy labels has been widely studied \cite{nt2019learning, zhong2019graph, dai2021nrgnn, li2021unified, fan2022partial}. We train the GCN with noisy labels for downstream inference. Our work could further contribute to this field as the Bayesian label transition model may recover the node label distribution using noisy labels with different noisy ratios.
{\bf 2) Uncertainty estimation.} Many researchers conduct studies on uncertainty estimation of node classifications using Bayesian approaches \cite{malinin2018predictive, zhang2019bayesian, zhao2020uncertainty, liu2020uncertainty, stadler2021graph}. Our work could dedicate to this field since the experiments demonstrate that the label inference can significantly decrease the uncertainty of node classifications.